%% file: tmm.tex
\documentclass[journal]{IEEEtran}

\usepackage{epsfig}
\usepackage{graphicx}
\usepackage{amsmath}
\usepackage{amssymb}

\usepackage{xcolor}
\usepackage{soul}
\usepackage[utf8]{inputenc}
\usepackage[pagebackref=true,breaklinks=true,letterpaper=true,colorlinks,bookmarks=false]{hyperref}

\usepackage[figuresright]{rotating}
\usepackage{textcomp,subfigure,algorithm,algorithmic,multirow,upgreek,tabularx}
\usepackage{graphics,gensymb}
\usepackage{bm,cases,threeparttable}

\usepackage{setspace}
\usepackage{pifont}

\graphicspath{{Figures/}}
\usepackage{booktabs}

\usepackage{amsfonts}
 
\usepackage{caption}
\usepackage{enumitem}
\setitemize{noitemsep,topsep=0pt,parsep=0pt,partopsep=0pt}

\usepackage{xcolor}
\usepackage{soul}
\usepackage[utf8]{inputenc}
\usepackage{mathrsfs}
\usepackage[figuresright]{rotating}
\usepackage{textcomp,subfigure,algorithm,algorithmic,multirow,upgreek,tabularx}
\usepackage{graphics,gensymb}
\usepackage{bm,cases,threeparttable}
\usepackage{bbm}
\usepackage{setspace}

\graphicspath{{Figures/}}
\usepackage{booktabs} 
\usepackage{amsfonts}

\ifCLASSINFOpdf
\else
\fi

\begin{document}
%
\title{
Hierarchical Cross-Attention Network for Virtual Try-On}


\author{\IEEEauthorblockN{Hao Tang,
		Bin Ren, Pingping Wu,
		Nicu Sebe
}
\thanks{
Hao Tang is with the School of Computer Science, Peking University, Beijing 100871, China. E-mail: haotang@pku.edu.cn 
\par Bin Ren and Nicu Sebe are with the Department of Information Engineering and Computer Science (DISI), University of Trento, Trento 38123, Italy. 
\par Pingping Wu is with the Jiangsu Key Laboratory of Public Project Audit, School of Engineering Audit, Nanjing Audit University, Nanjing 211815, China. Email: wupingping@nau.edu.cn
\par Corresponding author: Pingping Wu
}
}

\markboth{IEEE Transactions on Multimedia}%
{Shell \MakeLowercase{\textit{et al.}}: Bare Demo of IEEEtran.cls for IEEE Transactions on Magnetics Journals}
%

\IEEEtitleabstractindextext{%

\input{0abstract}

}

\maketitle

\IEEEdisplaynontitleabstractindextext

%
\IEEEpeerreviewmaketitle

\input{1introduction}
\input{2relatedwork}

\input{3method}
\input{4experiments}

\input{5conclusions}



\small
\bibliographystyle{IEEEtran}
\bibliography{ref}


\begin{IEEEbiography}[{\includegraphics[width=1in,height=1.25in,clip,keepaspectratio]{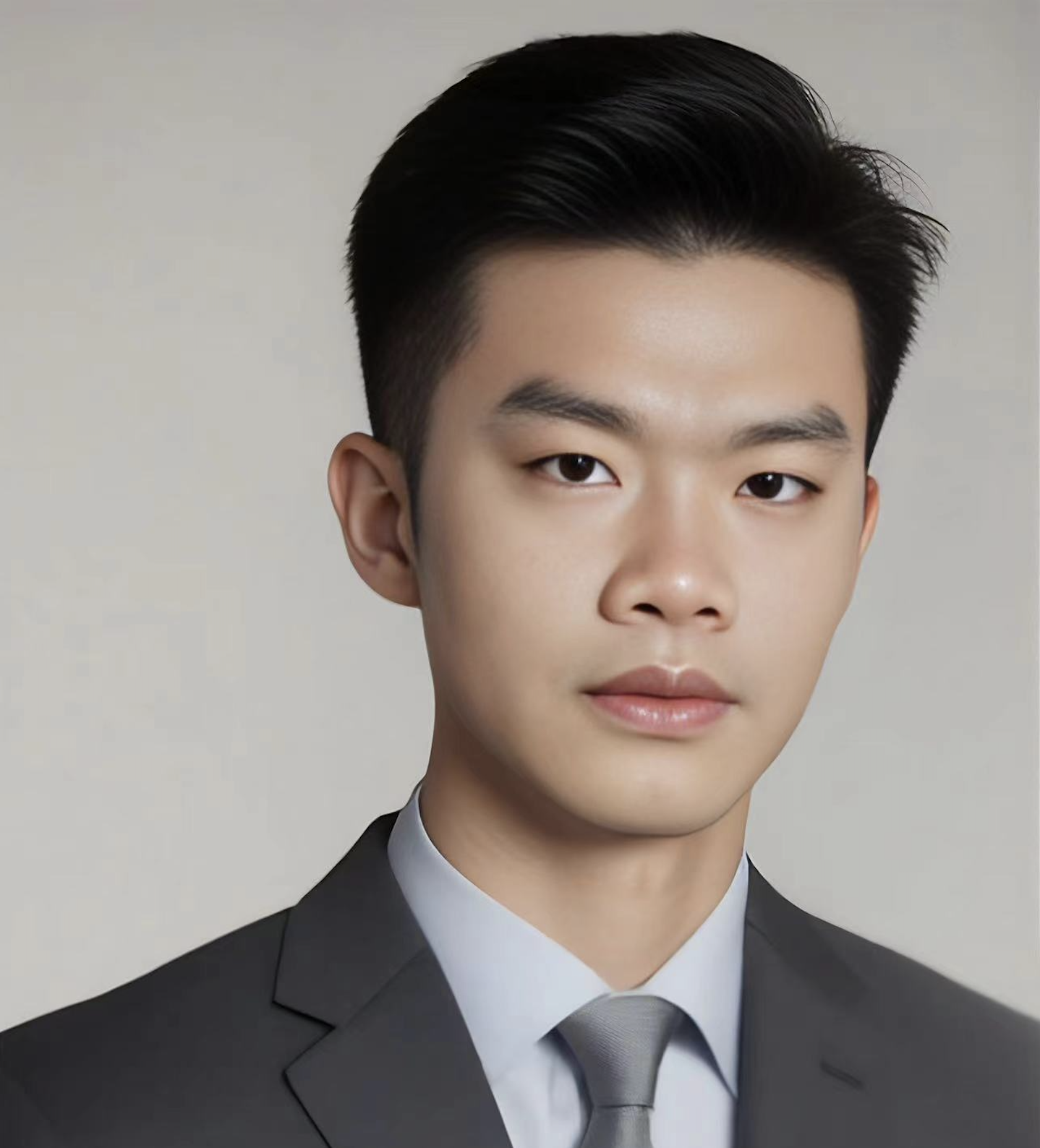}}]{Hao Tang}
is a tenure-track Assistant Professor at Peking University, China. Previously, he held postdoctoral positions at both CMU, USA, and ETH Zürich, Switzerland. He earned a master's degree from Peking University, China, and a Ph.D. from the University of Trento, Italy. He was a visiting Ph.D. student at the University of Oxford, UK, and an intern at IIAI, UAE. His research interests include AIGC, AI4Science, machine learning, computer vision, LLM, embodied AI, and their applications to scientific domains.
\end{IEEEbiography}

\begin{IEEEbiography}[{\includegraphics[width=1in,height=1.25in,clip,keepaspectratio]{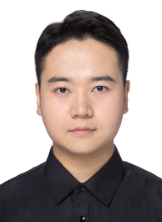}}]{Bin Ren}
is a Ph.D. candidate in the Department of Information Engineering and Computer Science at the University of Trento. He recieved the Master degree in computer application technology in 2021 at the School of Electronics and Computer Engineering, Peking University. He received the B.Eng. degree in the College of Mechanical and Electrical Engineering, Central South University. His research interests are deep learning, machine learning, and their applications to computer vision.
\end{IEEEbiography}

\begin{IEEEbiography}[{\includegraphics[width=1in,height=1.25in,clip,keepaspectratio]{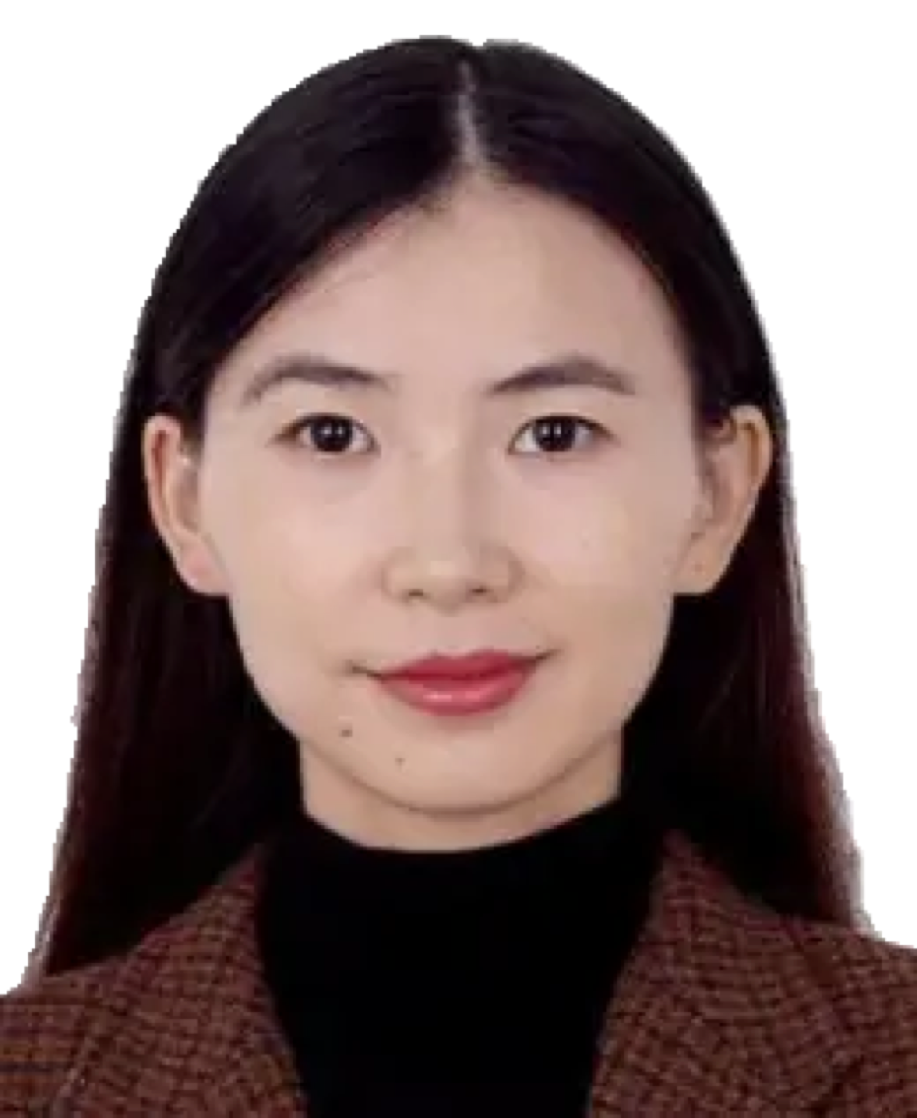}}]{Pingping Wu}
is currently an Associate Professor at Nanjing Audit University. She graduated from Peking University in 2016 with a major in Electronic Science and Technology, obtaining a doctoral degree in Science. Her research interests include computer vision, human-computer interaction, machine learning, and big data auditing. She has led and participated in multiple national and provincial research projects, and has published over 20 SCI papers in journals such as IEEE Transactions on Multimedia. \end{IEEEbiography}

\begin{IEEEbiography}[{\includegraphics[width=1in,height=1.25in,clip,keepaspectratio]{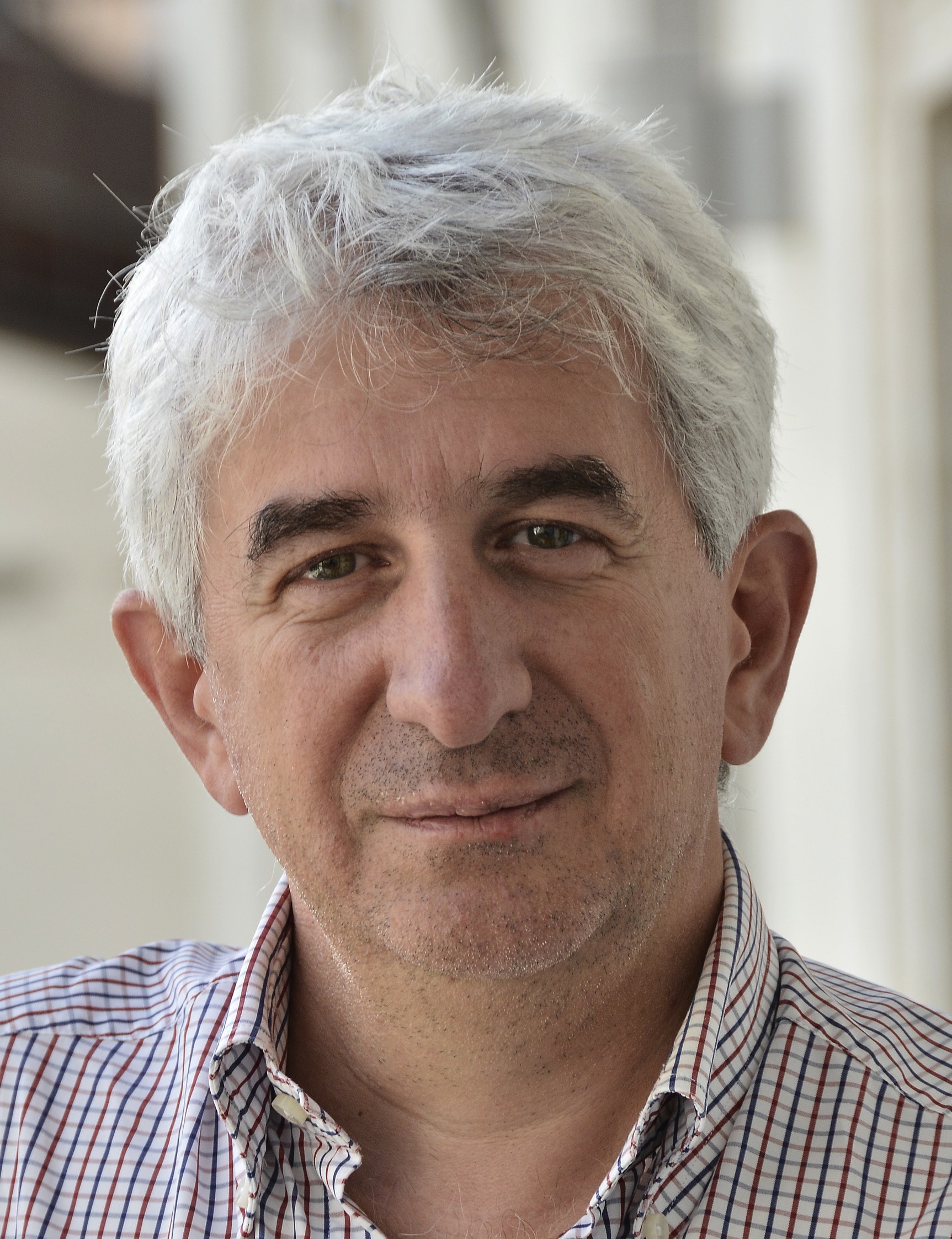}}]{Nicu Sebe}
is Professor in the University of Trento, Italy, where he is leading the research in the areas of multimedia analysis and human behavior understanding. He was the General Co-Chair of the IEEE FG 2008 and ACM Multimedia 2013.  He was a program chair of ACM Multimedia 2011 and 2007, ECCV 2016, ICCV 2017 and ICPR 2020.  He is a general chair of ACM Multimedia 2022 and a program chair of ECCV 2024. He is a fellow of IAPR.
\end{IEEEbiography}

%
%
%
%
%




\end{document}

%% file: 0abstract.tex
\begin{abstract}

In this paper, we present an innovative solution tailored for the intricate challenges of the virtual try-on task—our novel Hierarchical Cross-Attention Network, HCANet.
HCANet is meticulously crafted with two primary stages: geometric matching and try-on, each playing a crucial role in delivering realistic and visually convincing virtual try-on outcomes. A distinctive feature of HCANet is the incorporation of a novel Hierarchical Cross-Attention (HCA) block into both stages, enabling the effective capture of long-range correlations between individual and clothing modalities.
The HCA block functions as a cornerstone, enhancing the depth and robustness of the network. By adopting a hierarchical approach, it facilitates a nuanced representation of the interaction between the person and clothing, capturing intricate details essential for an authentic virtual try-on experience.
Our extensive set of experiments establishes the prowess of HCANet. The results showcase its cutting-edge performance across both objective quantitative metrics and subjective evaluations of visual realism. HCANet stands out as a state-of-the-art solution, demonstrating its capability to generate virtual try-on results that not only excel in accuracy but also satisfy subjective criteria of realism. This marks a significant step forward in advancing the field of virtual try-on technologies.

\end{abstract}

%% file: 1introduction.tex
\section{Introduction}

The virtual try-on task represents a major challenge that requires the generation of authentic, image-based virtual try-on results. Central to its objective is the accurate preservation of crucial attributes inherent to in-shop clothing, encompassing precise depiction of textures, shapes, and their appearance when adorned by consumers. This task has garnered substantial interest within the computer vision community because of its pronounced commercial advantages and practical implications. The applications of virtual try-on extend across diverse domains, ranging from online retail to the development of virtual fitting rooms, offering a multitude of transformative possibilities~\cite{han2018viton,jetchev2017conditional,wang2018toward,minar2020cp}.

The virtual try-on task has seen a diverse range of approaches, with promising results~\cite{ehara2006texture,brouet2012design,chen2016synthesizing,guan2012drape,sekine2014virtual,han2018viton,jetchev2017conditional,wang2018toward,minar2020cp,ge2021disentangled,yang2020towards,ge2021parser,lee2022towards,fenocchi2022dual,he2022style,huang2022towards,fele2022c,morelli2022dress,bai2022single,ren2023cloth}. These achievements underscore the ongoing innovation within this rapidly evolving field.
For example, VITON~\cite{han2018viton} employs a coarse-to-fine architecture to tackle the virtual try-on task. It initially computes a shape context~\cite{belongie2002shape} thin-plate spline (TPS) transformation~\cite{bookstein1989principal} to warp in-shop clothing onto the target individual. This is followed by blending techniques that contribute to the creation of a highly realistic appearance.
Another notable approach, PF-AFN~\cite{ge2021parser}, introduces a unique ``teacher-tutor-student'' knowledge distillation technique. This innovative method facilitates the generation of highly photorealistic images, setting a higher standard for quality and visual appeal in the realm of virtual try-on technology. These diverse methodologies highlight the ongoing exploration of advanced techniques and strategies to enhance the realism and quality of virtual try-on outcomes. The continuous evolution of the field is evident through these contributions, each pushing the boundaries of what is achievable in virtual try-on research.

Although studies like \cite{han2018viton,wang2018toward,minar2020cp,ge2021parser} have significantly advanced our understanding of virtual try-on tasks, they do come with certain limitations. Despite the progress achieved, issues persist, resulting in suboptimal results and visual anomalies in the generated images. 
These limitations underscore the complexity and challenges inherent in creating a genuinely realistic virtual try-on experience.
In particular, the existing virtual try-on methods mentioned above exhibit specific shortcomings in their approach. A prominent concern is their failure to account for the long-range global interactive correlations between the person's representation and the clothing's representation. This omission results in unsatisfactory results and inconsistent clothing-person pairings. In practical terms, this means that while the virtual image may resemble a person wearing clothing, the absence of a precise connection between the clothing and the individual's unique characteristics can lead to unnatural appearances. For instance, the clothing might appear as if it is floating over the body, or the texture might not align correctly with the body's contours. This challenge emphasizes the significance of a deep understanding of both geometry and appearance in the virtual try-on task. It extends beyond merely placing a piece of clothing on a digital representation of a person; it necessitates a thorough comprehension of how that clothing would realistically interact with the individual's distinct body shape, posture, and movement. This requires a deep understanding of how fabrics behave, how light interacts with surfaces, and how colors and textures should blend to achieve a genuinely lifelike appearance.
The identified issues within the existing methods pave the way for further research and development. By focusing on integrating long-range global interactive correlations, future virtual try-on technologies have the potential to attain a higher level of realism and accuracy. This could lead to more engaging and satisfying virtual shopping experiences, instilling greater confidence in online purchases for consumers, and potentially revolutionizing the online clothing shopping experience.

\begin{figure*}[t]
	\centering
	\includegraphics[width=0.9\linewidth]{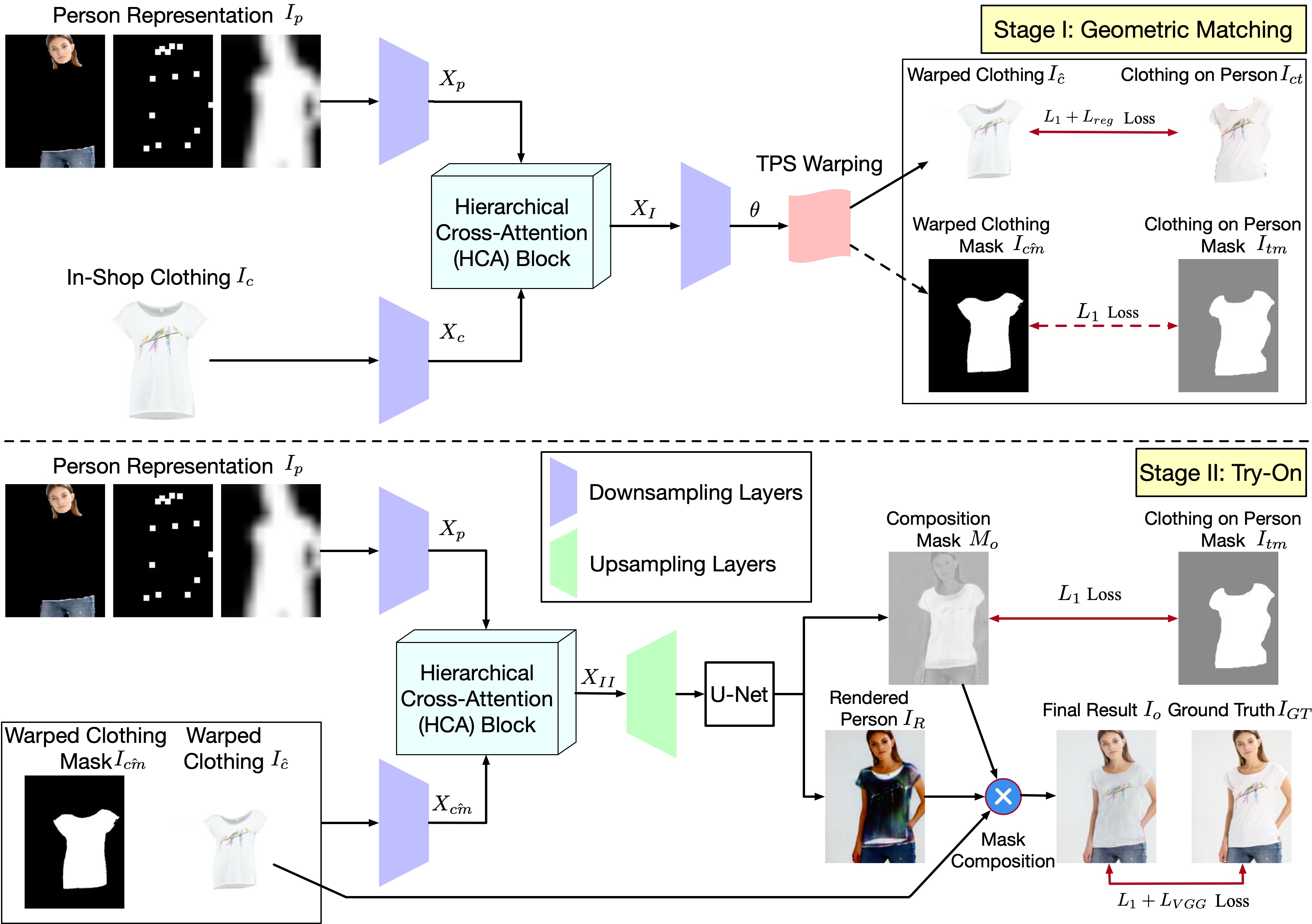}
	\caption{
	An overview of the proposed HCANet for the virtual try-on task reveals two key generative stages: geometric matching and try-on. Within both stages, we incorporate a novel hierarchical cross-attention (HCA) block designed to capture extensive correlations between the person and clothing modalities. The entire system operates in an end-to-end manner, fostering mutual enhancement between the stages to yield clothing images that are not only shape-consistent but also appearance-consistent. Conducting cross-attention operations twice in both stages of the HCA blocks offers several advantages. First, it allows for a more comprehensive integration of information from different modalities, such as RGB, joints, and binary masks, at multiple stages of the network. This can enhance the model's ability to capture complex relationships between these modalities and improve overall performance. Second, conducting cross-attention operations at multiple stages enables the model to refine its representations iteratively, facilitating more effective cross-modal interactions and feature fusion. This iterative refinement process can lead to better feature representations and ultimately improve the performance of the model.}
	\label{fig:tryon_method}
\end{figure*}

Expanding on this conceptual foundation, we present a novel contribution in the form of a novel hierarchical cross-attention network, denoted as HCANet.
HCANet, in its conceptualization, unfolds through two principal generative stages: geometric matching and try-on. To provide a comprehensive understanding of our proposed framework, we offer an illustrative overview in Figure~\ref{fig:tryon_method}.
Delving into the intricacies, the geometric matching stage serves the pivotal role of accurately aligning the in-shop clothing with the target person. This alignment is achieved through the application of a trainable thin-plate spline transformation, ensuring a precise and adaptable adjustment. Subsequently, the try-on stage takes the aligned clothing and person representations as input, orchestrating the creation of a pose-coherent image along with a composition mask.
The composition mask plays a critical role by incorporating information about the aligned clothing, thereby striking an optimal balance in the smoothness of the synthesized image. This strategic integration ensures that the virtual try-on outcomes exhibit a seamless blend between the person and the adorned clothing, resulting in a visually coherent and realistic representation.

The training process of the proposed HCANet unfolds in an end-to-end fashion, fostering a symbiotic relationship between the geometric matching and try-on stages. This holistic training approach enhances the synergy between the stages, facilitating the network's ability to learn and adapt cohesively, ultimately leading to improved performance and more faithful virtual try-on results.

Within both the geometric matching and try-on stages, we introduce a pioneering Hierarchical Cross-Attention (HCA) block to adeptly capture joint correlations between the person and clothing modalities. The structure of this HCA block is elucidated in Figure \ref{fig:pc} and comprises two distinct yet interconnected stages.
The first stage of the HCA block is dedicated to enhancing the fusion of diverse person representations through a cross-attention mechanism. This mechanism ensures that relevant features from different parts of the person's representation are effectively combined, contributing to a more comprehensive and nuanced understanding.
Simultaneously, the second stage of the HCA block focuses on the fusion of person and clothing representations, employing a parallel cross-attention approach. This dual-stage configuration is designed to create a hierarchical relationship between the person and clothing modalities, allowing for a synergistic fusion of information.
To further elaborate, the proposed HCA block computes the response at a specific position within one modality as a weighted sum of features from all positions within the other modality. This dynamic interplay fosters a more nuanced understanding of the correlation between the person and clothing, resulting in outcomes that are not only visually coherent but also remarkably photo-realistic.

Both qualitative and quantitative evaluations substantiate the superiority of HCANet over state-of-the-art methods. The proposed approach excels in terms of visual fidelity, ensuring a lifelike representation of the virtual try-on outcomes. Furthermore, it demonstrates superior alignment with the target person's poses, affirming its efficacy in delivering realistic and highly accurate virtual try-on results.

In summary, our paper makes several contributions to the field of virtual try-on:

\begin{itemize}
\item \textbf{Introduction of HCANet:} We present the innovative HCANet, a novel solution tailored specifically for the virtual try-on task. Comprising two distinct yet synergistic generation stages—geometric matching and try-on—HCANet establishes a comprehensive framework for generating realistic and visually convincing virtual try-on outcomes.

\item \textbf{Novel HCA Block:} Within HCANet, we introduce a novel Hierarchical Cross-Attention (HCA) block. This novel block is strategically designed to capture long-range global correlations between the person and clothing modalities. Deployed in both the geometric matching and try-on stages, the HCA block employs a hierarchical connection approach, enhancing the network's ability to discern intricate details and relationships, thereby contributing to more accurate and coherent virtual try-on results.

\item \textbf{SOTA Performance:} To substantiate the effectiveness of our proposed HCANet, we conducted extensive experiments on the challenging virtual try-on task. The results, derived from both quantitative metrics and qualitative assessments, unequivocally demonstrate the superior performance of HCANet. It not only outperforms existing methods, but also sets new benchmarks in terms of generating highly realistic and visually appealing virtual try-on outcomes.

\end{itemize}

%% file: 2relatedwork.tex
\section{Related Work}

\noindent \textbf{Virtual Try-On.} 
Virtual try-on, a highly popular task within the realm of fashion analysis, has garnered significant attention from the computer vision community due to its substantial commercial advantages and practical potential. Historically, this task was accomplished through computer graphics techniques, involving the creation of 3D models and the rendering of output images with precise control over geometric transformations or adherence to physical constraints~\cite{ehara2006texture,brouet2012design,chen2016synthesizing,guan2012drape,sekine2014virtual}. These methods, using 3D measurements or representations, could yield impressive results for virtual try-on (VTON).
Nevertheless, it is important to acknowledge the drawbacks associated with these techniques, including the need for extra 3D scanning equipment, significant computational resources, and intensive labor requirements.

Compared to 3D-based approaches, 2D image-based methods prove to be more practical for online shopping scenarios.
Jetchev and Bergmann~\cite{jetchev2017conditional} introduced a conditional analogy GAN for fashion item swapping solely using 2D images. However, their approach did not account for pose variation and required the pairing of in-shop clothing images with wearer images during inference, constraining its applicability.
VITON~\cite{han2018viton} addresses this limitation through a coarse-to-fine architecture. It initially computes a shape context~\cite{belongie2002shape} TPS transformation~\cite{bookstein1989principal} to warp in-shop clothing onto the target individual and subsequently blends the transformed garments onto the person.
Furthermore, both CP-VTON~\cite{wang2018toward} and CP-VTON+ \cite{minar2020cp} adopt similar two-stage frameworks to VITON but render the original TPS transformation learnable based on \cite{rocco2017convolutional}, using a convolutional geometric matcher. Although the try-on results appear more natural than those generated by VITON, they still exhibit significant artifacts when faced with heavy occlusions, intricate textures, or substantial transformations.
A more recent solution, ACGPN~\cite{yang2020towards}, aims to address these issues by adding a semantic generation module for spatial alignment. However, performance improvements are still in line with previous methods~\cite{han2018viton,wang2018toward,minar2020cp}, as they do not consider the latent long-range global interactive correlation between person and clothing modalities.
PF-AFN \cite{ge2021parser} introduces a unique ``teacher-tutor-student'' knowledge distillation technique, which excels at generating highly photorealistic images.
FS-VTON \cite{he2022style} proposes an innovative global appearance flow estimation model and leverages a StyleGAN-based architecture for appearance flow estimation, harnessing a global style vector to encode entire-image context, effectively addressing the challenges previously mentioned.
To overcome these limitations, we present a novel hierarchical cross-attention network (HCANet) explicitly designed to capture correlations, thereby producing more photorealistic and consistent results.

\noindent \textbf{Attention in Computer Vision.}
A myriad of studies have delved into the application of self- and cross-attention mechanisms across a spectrum of computer vision tasks. 
These tasks span 
image inpainting \cite{yu2018generative}, person re-identification \cite{wu2023learning,zhou2022cloth}, SAR target recognition \cite{ren2023transductive}, image denoising \cite{anwar2019real}, depth estimation \cite{xu2018structured, yang2021transformer}, image translation/generation \cite{tang2019multi, tang2021attentiongan, tang2019attention, tang2020dual, tang2019attribute, tang2020xinggan, tang2020edge, wu2022cross,tang2022multi}, layout generation \cite{tang2023graph}, cross-modal translation/analysis \cite{duan2021cascade, duan2021audio}, image retrieval \cite{ng2020solar}, object detection \cite{hu2022supervised}, and semantic segmentation \cite{ding2023few, ding2020lanet, yang2022continual}.
For example, Tang et al. \cite{tang2019multi} introduced SelectionGAN, a method that facilitates the generation of natural scene images from arbitrary viewpoints, given a scene image and a novel semantic map. Xu et al. \cite{xu2018structured} proposed a deep learning model for depth map calculation from still images, seamlessly integrating a front-end CNN and a multi-scale CRF, leveraging an attention mechanism for feature fusion. Yang et al. \cite{yang2021transformer} developed a novel decoder with gate-based attention mechanisms to preserve local-level detail capture in transformer-based networks. Wang et al. \cite{wang2020axial} introduced a position-sensitive axial-attention layer for image classification and dense prediction. Anwar et al. \cite{anwar2019real} utilized a residual on the residual structure and feature attention for image denoising.
In this paper, we contribute a novel approach, i.e., Hierarchical Cross-Attention (HCA), aimed at efficiently fusing person and clothing representations in a hierarchical manner for the intricate virtual try-on task. This HCA approach not only enhances the depth of the network but also yields more photorealistic results. Drawing inspiration from the success of attention mechanisms in a diverse range of computer vision applications, our HCA approach represents a significant stride forward in addressing the challenges posed by the virtual try-on task.

In summary, our work introduces two key innovations that differentiate it from CP-VTON+ \cite{minar2020cp}. Firstly, we present HCANet, a novel solution tailored specifically for the virtual try-on task, which encompasses two distinct yet synergistic generation stages: geometric matching and try-on. This comprehensive framework allows for the generation of realistic and visually convincing virtual try-on outcomes. Secondly, we introduce the novel HCA block within HCANet, strategically designed to capture long-range global correlations between the person and clothing modalities. Unlike CP-VTON+, our approach employs a hierarchical connection approach, enhancing the network's ability to discern intricate details and relationships, thus contributing to more accurate and coherent virtual try-on results.

%% file: 3method.tex
\begin{figure*}[t]
	\centering
	\includegraphics[width=1\linewidth]{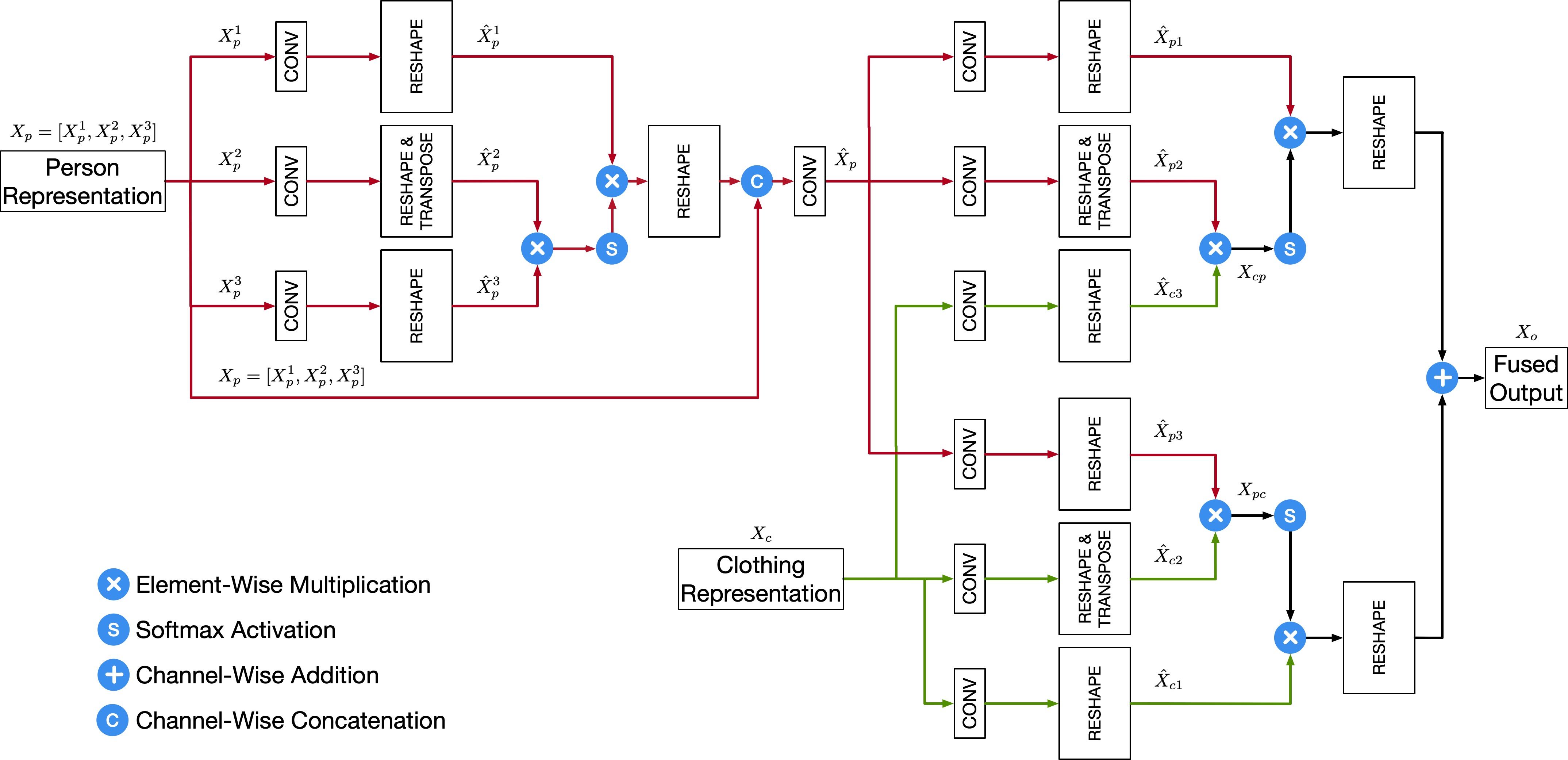}
	\caption{Structure of the proposed hierarchical cross-attention (HCA) block, which takes both the person representation and the clothing representation as inputs and produces the final fused interactive mutual correlation feature through a hierarchical cross-operation. 
Specifically, on
the left, cross-attention is performed on the person representations $X_p^1, X_p^2$, and $X_p^3$, giving the updated $\hat{X}_p$ as the output. On the right, cross-attention is computed between the clothing representation $X_c$ and the updated person representation $\hat{X}_p$. 
The symbols $\oplus$, $\otimes$, $\textcircled{s}$, and $\textcircled{c}$ denote element-wise addition, element-wise multiplication, Softmax activation, and channel-wise concatenation, respectively.}
	\label{fig:pc}
\end{figure*}

\section{The Proposed HCANet}

Many of the current image-based virtual try-on techniques suffer from suboptimal results as they often directly combine the representations of the person and in-shop clothing without taking into account their mutual interaction. 
Furthermore, conventional convolution operation struggles to establish long-range information, which is essential for producing globally consistent results.
To tackle these issues, we present an innovative solution: the HCANet (Hierarchical Cross-Attention Network) and introduce a novel hierarchical cross-attention (HCA) block for this specific task.
In the sections that follow, we will begin by offering an overview and presenting notations for our two-stage HCANet approach designed for virtual try-on (see Figure~\ref{fig:tryon_method}). This will be followed by a comprehensive exploration of the newly proposed HCA block (refer to Figure~\ref{fig:pc}) and the specific implementation details.

\subsection{Overview and Notations}
Consider a person image $I$ and an in-shop clothing image $I_{c}$. Our goal is to generate the image $I_{o}$, where a person $I_{o}$ wears the target clothing~$c$. 
We follow CP-VTON~\cite{wang2018toward} and CP-VTON+~\cite{minar2020cp} and adopt their basic structure in our proposed HCANet framework. 
Specifically, HCANet consists of two stages, i.e., geometric matching and try-on, as shown in Figure~\ref{fig:tryon_method}. 
The former aims at producing warped clothing and a warped mask according to a given person's pose and shape. 
The latter then utilizes these warped items, as well as the same person representation, to generate a final person image wearing the in-shop clothing.

In the first stage, HCANet takes person representation $I_{p}{=}[I_p^1, I_p^2, I_p^3]$ and in-shop clothing $I_{c}$ as inputs. Note that the person representation contains three components (as shown in Figure~\ref{fig:tryon_method}):
\begin{itemize}
	\item $I_p^1$: An 18-channel human pose heatmap with each channel corresponding to a human pose keypoint.
	\item $I_p^2$: A one-channel body shape feature map of a blurred binary mask that roughly covers different parts of the human body. 
	\item $I_p^3$: An RGB image that contains the reserved regions to maintain the identity of the person, including face, hair, etc.
\end{itemize}

Each input $I_p^1, I_p^2, I_p^3$, and $I_c$ is first passed through a downsampled feature extraction layer, which contains two-strided downsampling convolutional layers, followed by two one-strided ones, resulting in the feature map $X_p^1, X_p^2, X_p^3$, and $X_c$, respectively, where $X_p{=}[X_p^1, X_p^2, X_p^3]$ is the person representation and $X_c$ the in-shop clothing representation. 
Then, instead of directly using the matrix multiplication operation proposed by~\cite{rocco2017convolutional} such as CP-VTON~\cite{wang2018toward} and CP-VTON+~\cite{minar2020cp}, we replace it with our proposed HCA block to strengthen the ability to model global long-range relations. After that, the model can produce an improved correlation feature $X_{I}$.
A downsampling layer is then used to regress the parameter $\theta$. $\theta$ is used to warp the in-shop clothing $c$ using the on-body style $I_{\hat{c}}$ via the TPS warping module. Meanwhile, the relevant warped mask $I_{\hat{cm}}$ of $I_{\hat{c}}$ is also produced based on $\theta$. Note that this procedure is applied directly to the clothing mask $I_{cm}$, so we depict it in black dashes in Figure \ref{fig:tryon_method}.

In the second stage, we utilize the warped clothing $I_{\hat{c}}$ and the warped mask ${I_{\hat{cm}}}$ together with the person representation $I_{p}$ as input. Different from previous methods like CP-VTON or CP-VTON+, which directly concatenate the two warped items and person representation $I_{p}$, we again use the proposed HCA block here to model the interactive mutual relations between the warped clothing items and person features. Note that before the input data is passed through our proposed HCA block, we first employ downsampling convolutional layers to adjust the size of the input feature map and then apply an upsampling operation after the proposed block to return the feature $X_{II}$ to the original size. This ensures that the input of U-Net remains the same size as that for CP-VTON and CP-VTON+. Finally, a mask composition is constructed based on the warped clothing $I_{\hat{c}}$, the composition mask $M_{o}$, and the rendered person image $I_{R}$ (here $I_{R}$ refers to the human body outline generated by combining the person representation with the warped clothing mask, rather than the original person image). Consequently, the final result $I_{o}$ will be generated, and the overall mask composition operation can be described as follows:
\begin{equation}
\begin{aligned}
I_{o} & = M_{o} \times I_{\hat{c}} + (1 - M_{o}) \times I_{R}.
\label{eq:loss_1}
\end{aligned}
\end{equation}

Note that the first geometric matching stage is trained with sample triplets ($I_{p}$, $I_{c}$, and $I_{cm}$), while the second stage is trained with ($I_{p}$, $I_{\hat{c}}$, and $I_{\hat{cm}}$). 
Moreover, in the first matching stage, we adopt the same loss function used in CP-VTON+~\cite{minar2020cp} for fair comparison:
\begin{equation}
\begin{aligned}
L_{Matching} = \lambda_1 \cdot L_{1}(I_{\hat{c}}, I_{ct}) + \lambda_{reg} \cdot L_{reg},
\label{eq:loss_1}
\end{aligned}
\end{equation}
where $L_{1}$ indicates the pixel-wise $L1$ loss between the warped result $I_{\hat{c}}$ and the ground truth $I_{ct}$. $L_{reg}$ indicates the grid regularization loss formulated as follows:
\begin{equation}
\begin{aligned}
\begin{aligned}
L_{reg}\left(G_{x}, G_{y}\right)=& \sum_{i=-1,1} \sum_{x} \sum_{y}\left|G_{x}(x+i, y)-G_{x}(x, y)\right| \\
& \sum_{j=-1,1} \sum_{x} \sum_{y}\left|G_{x}(x, y+j)-G_{x}(x, y)\right|.
\end{aligned}
\label{eq:loss_1}
\end{aligned}
\end{equation}
where $G_{x}$ and $G_{y}$ indicate the grid coordinates of the generated images along the $x$ and the $y$ directions.
This is defined on the grid-related deformation item rather than directly on the TPS parameters for easy visualization and understanding.

In the second stage, the loss item is defined as follows:
\begin{equation}
\begin{aligned}
L_{Try-on} = \lambda_1 \cdot || I_{0} - I_{GT}||_{1} + \lambda_{vgg} \cdot L_{VGG} \\ + \lambda_{mask} \cdot ||M_{o} - I_{tm}||_{1}.
\label{eq:loss_1}
\end{aligned}
\end{equation}

The first term aims to minimize the discrepancy between the output $I_{o}$ and the ground truth $I_{GT}$. The second term, the VGG perceptual loss~\cite{johnson2016perceptual}, is widely used in image generation tasks, and the third term is used to encourage the composition mask $M_{o}$ to match the warped clothing mask $I_{tm}$ as much as possible.
Note that the VGG perceptual loss, derived from the feature representations of the VGG network, promotes perceptual similarity between the synthesized image and the ground truth by comparing high-level features extracted from both images. This perceptual similarity metric serves as a complement to traditional pixel-wise loss functions, such as $L_{1}$ loss, as it captures higher-level image characteristics, including texture and structural similarities. These aspects may not be fully captured by pixel-wise metrics alone.

\subsection{Hierarchical Cross-Attention Block}
The HCA block is designed to model the long-range global correlations between the person and clothing representations both in the first geometric matching stage and the second try-on stage.

The HCA block is given two input features, $X_p$ and $X_c$, which indicate the person and clothing representations, respectively, as shown in Figure~\ref{fig:pc}. 
Our goal is to obtain the final fused output $X_{o}$, which contains the mutual correlations between the person and clothing features. 

Specifically, the proposed HCA block contains two stages, which are cross-attention for person representations, 
and cross-attention for person and clothing representations. 
The two types of cross-attention are
carried out in sequence, and they together align and fuse the person representation $X_p$ and the clothing representation $X_c$. 
We describe them in the following two subsections.

\noindent\textbf{Cross-Attention for Person Representations.} We first feed $X_{p}{=}[X_p^1, X_p^2, X_p^3]$ into three different convolutional layers to generate three different intermediate features, resulting in three kinds of features, named $\tilde{X}_{p}^1$, $\tilde{X}_{p}^2$, and $\tilde{X}_{p}^3$, respectively. We then reshape and transpose $\tilde{X}_{p}^2$ from ${\mathbb{R}^{c \times h \times w}}$ to $\hat{X}_{p}^2 {\in}{\mathbb{R}^{n \times c}}$. At the same time, we also reshape the feature $\tilde{X}_{p}^3$ from ${\mathbb{R}^{c \times h \times w}}$ to $\hat{X}_{p}^3{\in}{\mathbb{R}^{c \times n}}$. Note that here $n{=}hw$ is the number of pixels. After that, we perform a matrix multiplication between $\hat{X}_{p}^{2}$ and $\hat{X}_{p}^{3}$ and apply a Softmax layer to produce the person-based interactive correlation matrix $X_{pp}{\in}{\mathbb{R}^{hw \times hw}}$:
\begin{equation}
\begin{aligned}
X_{pp}^{ji} = \frac{{\rm exp} (\hat{X}_{pi}^{2} \hat{X}_{pj}^{3})}{\sum_{i=1}^n {\rm exp}(\hat{X}_{pi}^{2}  \hat{X}_{pj}^{3})},\\
\end{aligned}
\end{equation}
where $X_{pp}^{ji}$ measures the impact of the $i$-th position of the feature $\hat{X}_{p}^{2}$ on the $j$-th position of the feature $\hat{X}_{p}^{3}$. This cross strategy ensures that a global pairwise similarity correlation can be built between different person representations.

Consequently, the refined person representation output $\hat{X}_{p}$ is produced as follows:
\begin{equation}
\begin{aligned}
\hat{X}_p = {\rm Conv}\{{\rm Concat}[\sum_{i=1}^{n}(X_{pp}^{ji}  \hat{X}_{pi}^{1}), X_p^1, X_p^2, X_p^3]\},
\end{aligned}
\end{equation}
where ${\rm Concat}(\cdot)$ and ${\rm Conv}(\cdot)$ denote the channel-wise concatenation and convolutional layer, respectively.

\noindent\textbf{Cross-Attention for Person and Clothing Representations.}
We first feed both the $\hat{X}_{p}$ and $X_{c}$ features into six different convolutional layers to generate six different intermediate features. Specifically, from the person representation, we can obtain three kinds of features, named $\tilde{X}_{p1}$, $\tilde{X}_{p2}$, and $\tilde{X}_{p3}$, which are depicted in red in Figure~\ref{fig:pc}. Similarly, the clothing representation also generates three kinds of features, named $\tilde{X}_{c1}$, $\tilde{X}_{c2}$, and $\tilde{X}_{c3}$, shown in green in Figure~\ref{fig:pc}. We first reshape and transpose $\tilde{X}_{p2}$ from ${\mathbb{R}^{c \times h \times w}}$ to $\hat{X}_{p2}{\in}{\mathbb{R}^{n \times c}}$. At the same time, we also reshape the clothing feature $\tilde{X}_{c3}$ from ${\mathbb{R}^{c \times h \times w}}$ to $\hat{X}_{c3}{\in}{\mathbb{R}^{c \times n}}$. Note that here $n{=}hw$ is the number of pixels. After that, we perform a matrix multiplication between $\hat{X}_{p2}$ and $\hat{X}_{c3}$, and apply a Softmax layer to produce the clothing-guided person-based interactive correlation matrix $X_{cp}{\in}{\mathbb{R}^{hw \times hw}}$:

\begin{equation}
\begin{aligned}
X_{cp}^{ji} = \frac{{\rm exp} (\hat{X}_{c3}^{i}  \hat{X}_{p2}^{j})}{\sum_{i=1}^n {\rm exp}(\hat{X}_{c3}^{i}  \hat{X}_{p2}^{j})},\\
\end{aligned}
\end{equation}
where $X_{cp}^{ji}$ measures the impact of the $i$-th position of clothing feature $\hat{X}_{c3}$ on the $j$-th position of the person feature $\hat{X}_{p2}$. This cross strategy ensures that a global pairwise similarity correlation can be built between the person and clothing representations.
Consequently, the refined person representation output $X_{cp}^{o}$ is produced as follows:
\begin{equation}
\begin{aligned}
X_{cp}^{o} = \sum_{i=1}^{n}(X_{cp}^{ji}  \hat{X}_{p1}^{i}).
\end{aligned}
\end{equation}

Following the same pipeline for obtaining the clothing-guided person-based interactive correlation matrix $X_{cp}$ and refined person representation, we can also obtain the person-guided clothing-based interactive correlation matrix $X_{pc}$ and refined clothing representation $X_{pc}^{o}$ from $\hat{X}_{c1}$, $\hat{X}_{c2}$, and $\hat{X}_{p3}$, as follows:
\begin{equation}
\begin{aligned}
X_{pc}^{ji} & = \frac{{\rm exp} (\hat{X}_{p3}^{i}  \hat{X}_{c2}^{j})}{\sum_{i=1}^n {\rm exp}(\hat{X}_{p3}^{i}  \hat{X}_{c2}^{j})},\\
X_{pc}^{o} & = \sum_{i=1}^{n}(X_{pc}^{ji}  \hat{X}_{c1}^{i}).
\end{aligned}
\end{equation}

Then, the final fused output of our HCA block can be computed by simply combining the reshaped versions of $X_{cp}^{o}$ and $X_{pc}^{o}$:
\begin{equation}
\begin{aligned}
X_{o} = X_{cp}^{o} + X_{pc}^{o}.
\end{aligned}
\end{equation}

Very different from previous methods like CP-VTON or CP-VTON+, which directly concatenate the original clothing and person representations together, our proposed HCA block takes the interactive relation between the inputs into consideration. Moreover, our cross strategy is strong in modeling long-range global dependencies, producing more photo-realistic try-on images.

\subsection{Implementation Details}
We use similar settings as CP-VTON~\cite{wang2018toward} and CP-VTON+~\cite{minar2020cp} in our experiments. 
Specifically, we set $\lambda_{1} {=} \lambda_{vgg} {=} \lambda_{mask} {=} 1$ during the training. Both stages are trained for 200K steps with batch size 4. Moreover, for the Adam optimizer, $\beta_{1}$ and $\beta_{2}$ are set to 0.5 and 0.999, respectively. The learning rate is fixed at 0.0001 for the first 100K steps and then linearly decayed to 0 over the remaining steps. All input images are resized to $256 {\times} 192$, and the output images have the same resolution.

%% file: 4experiments.tex
\section{Experiments}

This section will provide experimental results and analyses of HCANet for the virtual try-on task.

\begin{figure*}[!t]
	\centering
	\includegraphics[width=1\linewidth]{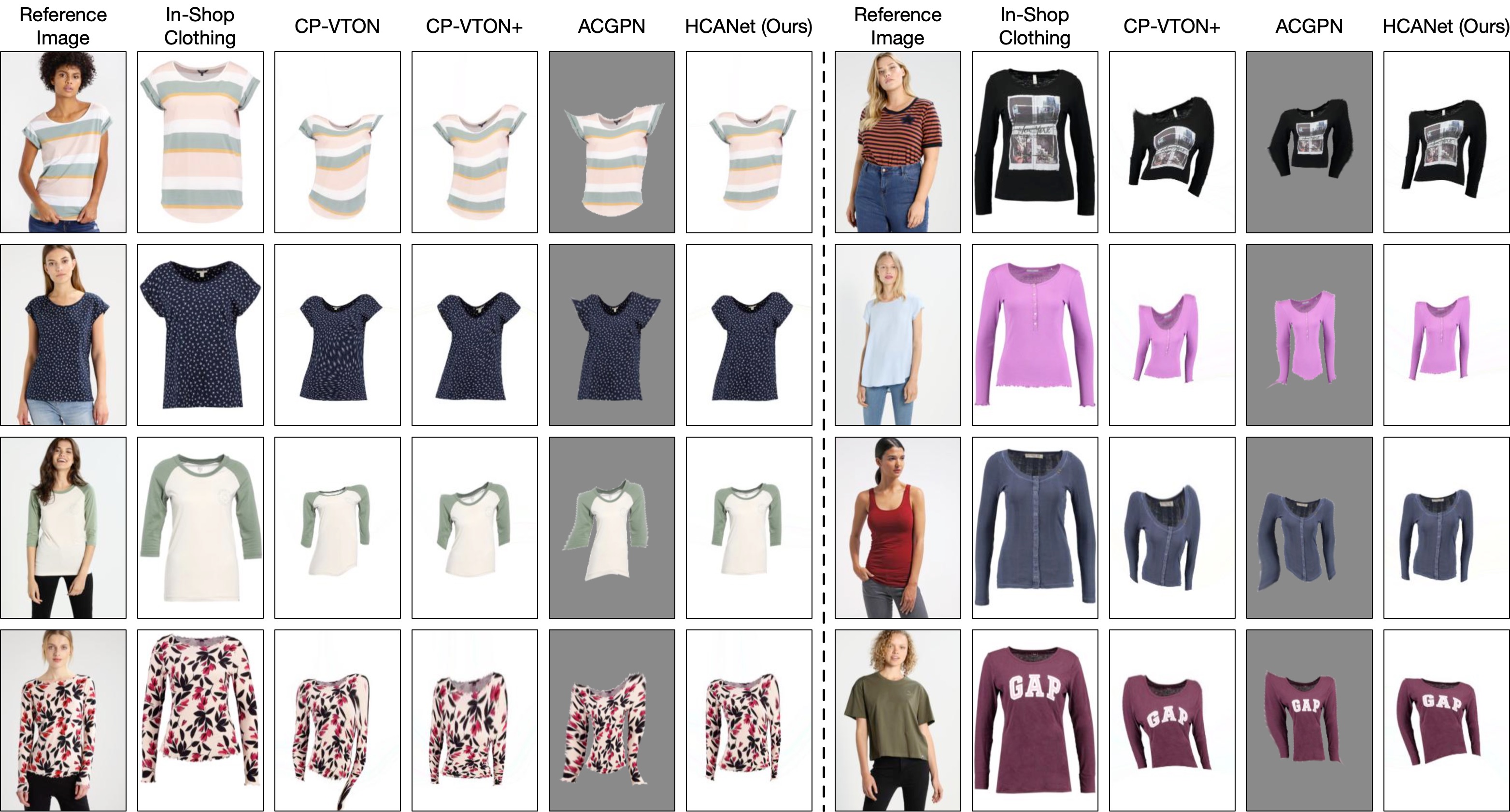}
	\caption{Qualitative comparisons of the clothes warped by CP-VTON~\cite{wang2018toward}, CP-VTON+~\cite{minar2020cp}, and ACGPN~\cite{yang2020towards} in the first geometric matching stage. To the left of the dashed line are same-pair (retry-on) cases, while to the right are the different-pair cases.}
	\label{fig:warp_result}
\end{figure*}

\begin{table*}[!t]
	\centering
	\caption{Quantitative results of virtual try-on, where input pairs are same-clothing for IoU, SSIM, and LPIPS, and different-clothed for IS.}
	\begin{tabular}{rcccc} \toprule
		Method  & IoU $\uparrow$ & SSIM $\uparrow$ & LPIPS $\downarrow$ & IS (mean $\pm$ std.) $\uparrow$  \\ \midrule	
		CP-VTON~\cite{wang2018toward}  & 0.7898 & 0.800 & 0.126 & 2.7809 $\pm$ 0.0594 \\ 
		CP-VTON+~\cite{minar2020cp}  & 0.8425 & 0.817 & 0.117 & 3.1048 $\pm$ 0.1068  \\ 
		ACGPN~\cite{yang2020towards} & - & 0.846 & 0.121 & 2.829 \\ 
            CIT \cite{ren2023cloth} & - & 0.827 & 0.115 & 3.060 \\
		HCANet (Ours) & \textbf{0.8662} & \textbf{0.859} & \textbf{0.102} & \textbf{3.2483 $\pm$ 0.0865} \\
		\bottomrule	
	\end{tabular}
	\label{tab:sota}
\end{table*}

\subsection{Experimental Setups}

\noindent\textbf{Datasets.} Our study adheres to the methodology employed by both CP-VTON~\cite{wang2018toward} and CP-VTON+ \cite{minar2020cp} for experimentation. We select the dataset assembled by Han et al. \cite{han2018viton}, a widely recognized and meticulously curated collection in the field. This dataset choice not only provides a robust foundation but also aligns our work with prior significant contributions in the realm of virtual try-on technology.
To respect copyright restrictions, we employ a reorganized version of the dataset proposed in \cite{minar2020cp} for a fair comparison. Complying with legal considerations is a fundamental aspect of research, upholding the integrity and ethical standards of our study.
The dataset in use encompasses approximately 19,000 pairs of front-view images featuring women and top clothing. This extensive assortment offers a diverse and rich sample, representing a wide array of styles, fabrics, and fits. The diversity within the dataset enhances the resilience and real-world applicability of our study. Within this dataset, 16,253 pairs have undergone meticulous curation. The cleaning process involves refining and rectifying the images to ensure they meet stringent standards, eliminating any inconsistencies or errors that could potentially bias the results.
This curated data is then partitioned into a training set and a validation set, consisting of 14,221 and 2,032 pairs, respectively. This partition adheres to standard machine learning practices, enabling an unbiased assessment of the algorithm's performance.
A notable feature of this dataset is that, in the training set, the target clothing and the clothing worn by the wearer are identical. This design choice ensures a consistent and controlled environment for model training. In contrast, the test set introduces variability by featuring distinct target clothing and attire worn by the wearer. This design decision introduces a more complex and realistic challenge for evaluation, simulating real-world scenarios where users would seek to visualize different clothing on the same person.

\begin{figure*}[!t]
	\centering
	\includegraphics[width=1\linewidth]{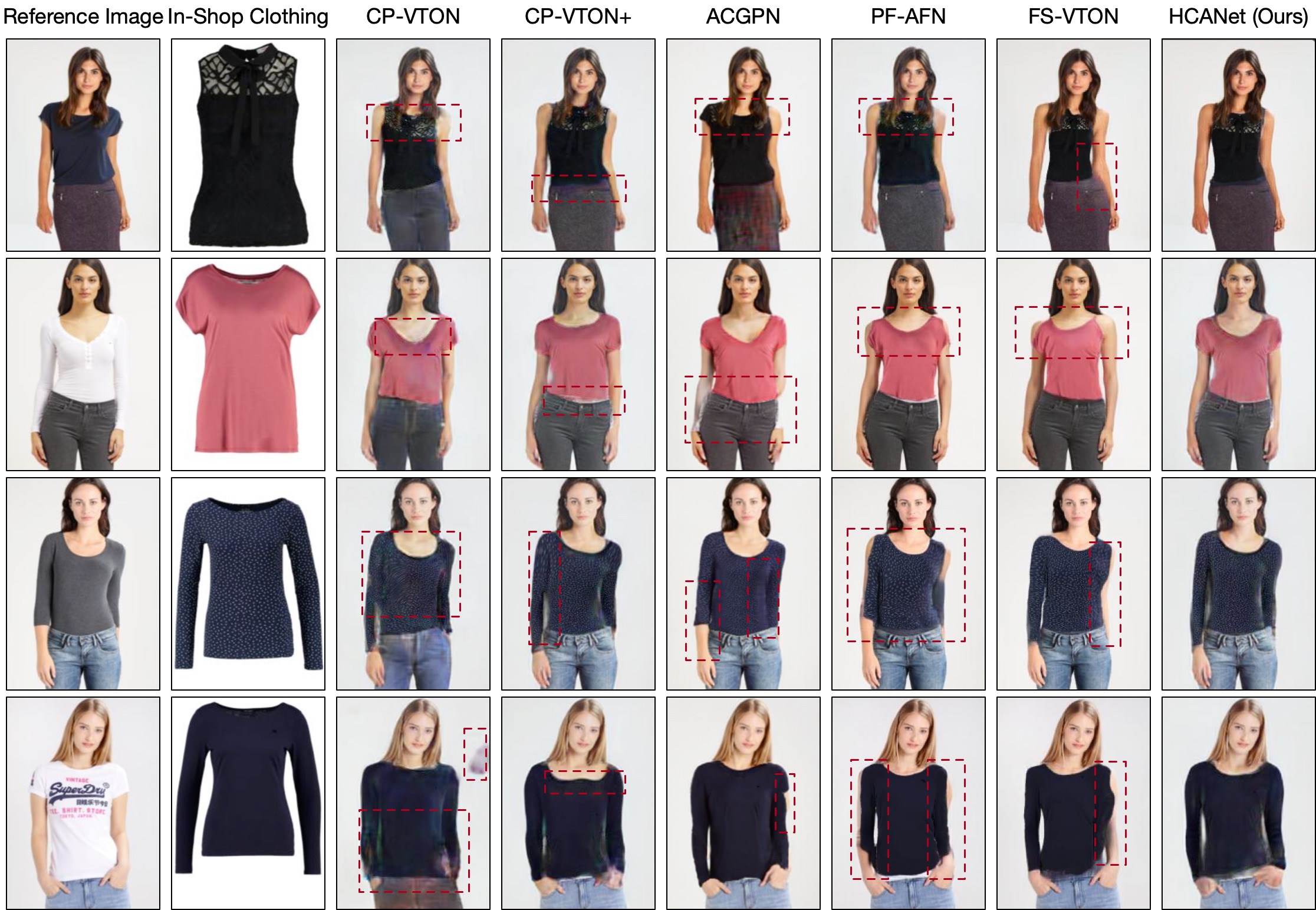}
	\caption{Qualitative comparison with CP-VTON~\cite{wang2018toward}, CP-VTON+~\cite{minar2020cp}, ACGPN~\cite{yang2020towards}, PF-AFN \cite{ge2021parser}, FS-VTON \cite{he2022style} in the second try-on stage.}
	\label{fig:tryon_sota}
\end{figure*}

\noindent\textbf{Evaluation Metrics.}
In line with CP-VTON+\cite{minar2020cp}, we employ the intersection over union (IoU), SSIM \cite{wang2004image}, and learned perceptual image patch similarity (LPIPS) \cite{zhang2018unreasonable} metrics for assessing the performance in cases of retry-on with the same clothing (ground truth available) during both the initial geometric matching stage and the subsequent try-on stage. To be more specific, we use the original target human image as the reference for SSIM and LPIPS, while the parsed segmentation area of the current upper clothing serves as the IoU reference. In scenarios involving the try-on of different clothing (lacking ground truth), we opt for the Inception Score (IS) \cite{salimans2016improved} as our evaluation metric. Furthermore, in keeping with FS-VTON \cite{he2022style}, we also include the Fr'echet Inception Distance (FID) \cite{heusel2017gans} metric in our evaluation.

\subsection{Experimental Results}

\noindent\textbf{Quantitative Comparisons.}
In Table \ref{tab:sota}, we present the results for IoU, SSIM, LPIPS, and IS, comparing them with CP-VTON~\cite{wang2018toward}, CP-VTON+ \cite{minar2020cp}, ACGPN \cite{yang2020towards}, and CIT \cite{ren2023cloth}. These evaluations are performed in the context of the first geometric matching stage and the second try-on stage. Specifically, IoU assesses the quality of the warped mask during the first geometric matching stage with same-pair test samples, while SSIM, LPIPS, and IS evaluate the second try-on stage. It is important to note that SSIM and LPIPS are used for assessing same-pair images (retry-on), while IS is employed for evaluating unpaired try-on results. 
Table \ref{tab:sota} showcases the superiority of the proposed HCANet, with the highest scores on all four metrics compared to other state-of-the-art methods. These results validate the effectiveness of our approach.

Additionally, we report the FID results in Table \ref{tab:fid}. It is evident that HCANet outperforms existing state-of-the-art methods, including VTON \cite{han2018viton}, CP-VTON~\cite{wang2018toward}, CP-VTON+\cite{minar2020cp}, ClothFlow \cite{han2019clothflow}, ACGPN\cite{yang2020towards}, DCTON \cite{ge2021disentangled}, PF-AFN \cite{ge2021parser}, ZFlow \cite{chopra2021zflow}, USC-PFN \cite{du2024greatness}, DOC-VTON \cite{yang2023occlumix}, and FS-VTON \cite{he2022style}. Significantly, even compared to the already low FID score of 8.89 achieved by FS-VTON, our method further reduces it to 8.25, establishing a new state-of-the-art performance in this challenging task.

\begin{table}[!t]
	\centering
	\caption{Quantitative FID results of virtual try-on.}
		\begin{tabular}{rc} \toprule
			Method  & FID $\downarrow$ \\ \midrule
                VTON \cite{han2018viton}          & 55.71 \\
                CP-VTON~\cite{wang2018toward}     & 24.45 \\
                CP-VTON+~\cite{minar2020cp}       & 21.04  \\
                ClothFlow \cite{han2019clothflow} & 14.43 \\
                ACGPN~\cite{yang2020towards}      & 16.64 \\ 
                DCTON \cite{ge2021disentangled}   & 14.82\\
                PF-AFN \cite{ge2021parser}        & 10.09\\ 
                ZFlow \cite{chopra2021zflow}      & 15.17\\
                USC-PFN \cite{du2024greatness} & 10.47 \\
                DOC-VTON \cite{yang2023occlumix} & 9.54 \\
                FS-VTON \cite{he2022style}        & 8.89\\ \hline
			HCANet (Ours)                     & \textbf{8.25} \\
			\bottomrule	
	\end{tabular}
	\label{tab:fid}
\end{table}

\begin{figure}[!t]
	\centering
	\includegraphics[width=0.8\linewidth]{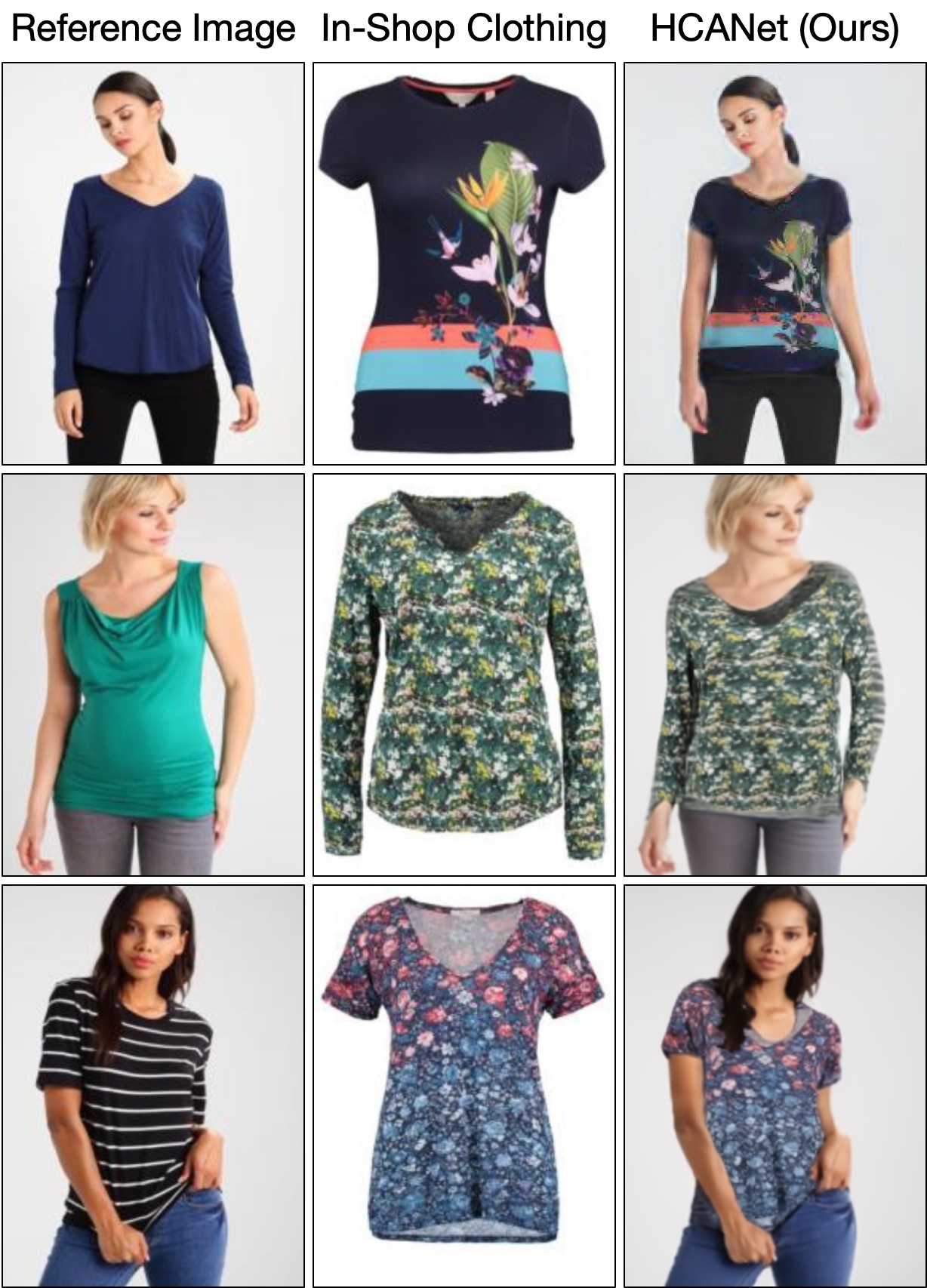}
	\caption{Try-on results of clothes with complex textures.}
	\label{fig:tryon_sota_more}
\end{figure}

\noindent\textbf{Qualitative Comparisons.}
To provide a more comprehensive view of HCANet's performance in virtual try-on, we offer visualization comparisons from both stages, showcasing the warped clothing in Figure~\ref{fig:warp_result} and the final try-on person images in Figures~\ref{fig:tryon_sota} and \ref{fig:tryon_sota_more}.
In Figure~\ref{fig:warp_result}, we present the results for warped clothing in both retry-on (same-pair) and try-on (different-pair) scenarios. It is evident that HCANet excels in generating sharper and more realistic warped clothing when compared to other state-of-the-art methods. For instance, in qualitative comparisons, our method consistently outperforms CP-VTON+, particularly in texture-rich cases like line stripes in the first row and intricate clothing patterns in the last row of the same-pair cases. CP-VTON+, in contrast, struggles to warp in-shop clothing articles in a reasonable direction.

Figure~\ref{fig:tryon_sota} displays the final results of different-pair try-on scenarios, further highlighting HCANet's superiority over other state-of-the-art methods. Our approach excels in preserving the original clothing texture and pattern, resulting in more realistic and natural generated images. In contrast, other methods exhibit numerous artifacts, including unrealistic outcomes for unique clothing (the first row), excessively warped clothing textures (the second and third rows), and irregular clothing patterns (other cases). These results reinforce the superiority of our method.

However, upon close inspection, we observe that the generated depiction of the person's hands is notably distorted, and there is a conspicuous presence of white light and shadow at the junction where the clothes meet the pants. These issues indicate that while the overall generation process is functioning as intended, there are specific areas where the model is struggling to produce accurate and realistic outputs.\\
\indent
The distortion of the generated hands likely stems from the absence of detailed pose information specifically related to the hands in the input data. In image generation tasks, particularly those involving complex human poses, the accurate representation of the hands is challenging due to their intricate structure and the wide range of possible positions they can assume. Hands are highly expressive and can adopt various subtle poses that convey different meanings or actions, making them one of the most difficult parts of the body to model accurately in generative tasks. \\
\indent
In our case, the lack of a detailed pose representation means that the model has to infer the hand positions based on incomplete or generalized information. This can lead to inaccurate or unnatural hand shapes, such as fingers being incorrectly positioned, unnatural bending, or even merging with other parts of the body or clothing. The model may rely too heavily on prior knowledge or assumptions about hand shapes, resulting in distortions when the actual hand pose deviates from these assumptions.\\
\indent
Providing detailed pose information for the hands could significantly alleviate this problem by guiding the model to generate hands that are anatomically correct and visually consistent with the rest of the body. This could involve incorporating additional data such as hand keypoints or pose annotations that specifically capture the orientation and positioning of the fingers and palms. With this information, the model can more accurately render the hands, reducing the likelihood of distortion and ensuring that the generated images are more realistic and coherent. \\
\indent
In summary, the distortion of the hands in the generated images highlights the importance of providing comprehensive pose information, especially for complex and expressive parts of the body like the hands. By enhancing the input data with detailed hand poses, we can improve the quality of the generated outputs and produce images that are more true to life.

Meanwhile, the presence of white light and shadow at the junction of clothes and pants in the try-on results may be attributed to two factors: (1) Lighting inconsistency: variations in lighting conditions between the original image and the virtual try-on environment can result in discrepancies in light and shadow rendering; (2) Resolution and quality of input images: low-resolution or poorly captured input images may exacerbate issues with light and shadow rendering in the generated try-on results.

In Figure \ref{fig:tryon_sota_more}, we provide more try-on results of clothes with complex textures. These results provide insight into the effectiveness of our approach in accurately rendering the details and nuances of clothing textures, which are crucial to enhancing the realism of virtual try-on experiences.

\noindent\textbf{User Study.}
To further assess the performance of HCANet and compare it to other leading methods, we conducted two user studies.
In the first setting, we randomly selected 100 sets of reference images and target clothing images from the test dataset. Participants were presented with the reference images and target clothing and asked to choose the best outputs from our model and the baseline methods (CP-VTON \cite{wang2018toward}, CP-VTON+~\cite{minar2020cp}, ACGPN \cite{yang2020towards}, PF-AFN \cite{ge2021parser}, FS-VTON \cite{he2022style}). Their choices were based on two questions: the first question inquired about the most photo-realistic image, and the second question focused on which image preserved the most details of the target clothing.

\begin{table}[!t] \small
	\centering
	\caption{User study of setting 1 ($\%$).
		``Image Realism'' denotes the question ``Which image is the most photo-realistic?'', and ``Clothing Consistency'' denotes the question ``Which image preserves the most details of the target clothing?''.}
	\resizebox{1\linewidth}{!}{%
		\begin{tabular}{rcc} \toprule
			Method  & Image Realism $\uparrow$ & Clothing Consistency $\uparrow$ \\ \midrule	
			CP-VTON~\cite{wang2018toward}& 4.6 & 4.1   \\ 
			CP-VTON+~\cite{minar2020cp}  & 8.4 & 10.2 \\
			ACGPN~\cite{yang2020towards} & 16.8 & 15.9 \\
			PF-AFN  \cite{ge2021parser}  & 19.7 & 17.5 \\
			FS-VTON \cite{he2022style}   & 23.3 & 22.5 \\ \hline
			HCANet (Ours)               & \textbf{27.2} & \textbf{29.8} 
			\\
			\bottomrule	
	\end{tabular}}
	\label{tab:sota_us}
\end{table}

\begin{table}[!t] 
	\centering
	\caption{User study of setting 2 ($\%$).
		The preference rate comparing other models against our
		model (other models/our model) in human evaluation.}
	\resizebox{0.6\linewidth}{!}{%
		\begin{tabular}{rc} \toprule
			Compared Method  & Preference Rate $\uparrow$ \\ \midrule	
			CP-VTON~\cite{wang2018toward} & 8.3 / 91.7  \\ 
			CP-VTON+~\cite{minar2020cp}   & 10.2 / 89.8 \\
			ACGPN~\cite{yang2020towards}  & 16.6 / 83.4 \\
			PF-AFN  \cite{ge2021parser}   & 30.7 / 69.3 \\
			FS-VTON \cite{he2022style}    & 44.6 / 55.4 
			\\ \bottomrule	
	\end{tabular}}
	\label{tab:sota_us2}
\end{table}

For setting 2, we adopted the approach outlined in FS-VTON \cite{he2022style}, presenting users with an input person image, a clothing image, and the generated try-on image from two models (CP-VTON \cite{wang2018toward}, CP-VTON+~\cite{minar2020cp}, ACGPN \cite{yang2020towards}, PF-AFN \cite{ge2021parser}, FS-VTON \cite{he2022style}). Users were asked to vote on the superior generated try-on image.
Each user was randomly assigned 100 images to compare two models.

In total, 30 users participated in both evaluation for all model comparisons. Our user study subjects consisted of students and teachers between 25 and 50 years of age. We ensure a balanced representation of gender, with 16 male participants and 14 female participants. All participants have a certain level of understanding and experience with virtual try-on technologies. To mitigate potential biases, we employ strategies such as random selection to ensure equal representation across age groups and genders. In addition, we implement measures to minimize response biases, including the use of anonymous surveys. By including this information, we aim to improve the transparency and reliability of our user study findings.

The results from setting 1 are summarized in Table~\ref{tab:sota_us}, where it is clear that HCANet outperforms other leading methods for both questions, reaffirming the superior photo-realism of the images generated by our model.
It is noteworthy that some methods perform well in the ``Image Realism'' metric while significantly underperforming in the ``Clothing Consistency'' metric. This discrepancy can be attributed to the differing focus of these two evaluation metrics. While ``Image Realism'' primarily assesses the visual fidelity and naturalness of the synthesized images, ``Clothing Consistency'' emphasizes the accuracy of how the clothing interacts with the person's body and conforms to its shape and contours. Therefore, a method may excel in generating realistic-looking images but may struggle to maintain consistency in clothing fit and appearance. Conversely, a method that prioritizes clothing consistency may sacrifice some aspects of image realism to ensure better adherence to the clothing's structure and details. By recognizing the different objectives of each metric, we gain a deeper understanding of the trade-offs involved in virtual try-on methods and can better interpret their performance across different evaluation criteria.
The results from setting 2 can be found in Table~\ref{tab:sota_us2}. These results align with those in Tables \ref{tab:sota}, \ref{tab:fid}, and \ref{tab:sota_us}, confirming our method's consistent superiority with a preference rate exceeding 10\% when compared to all the models.

\subsection{Ablation Study}
\noindent\textbf{Variants of HCANet.}
To assess the effectiveness of the proposed HCANet in the virtual try-on task, we conducted five ablation experiments, labeled as B0, B1, B2, B3, and B4, as detailed in Table~\ref{tab:ablation}.
B0 utilizes CP-VTON+~\cite{minar2020cp} as the baseline.
B1 incorporates the HCA block solely in the first geometric matching stage.
B2 integrates the HCA block exclusively in the second try-on stage.
B3 employs the HCA block in both stages.
B4 represents the full model, an extension of B3, with the inclusion of an additional $L_{1}$ loss between the warped clothing mask $I_{\hat{cm}}$ and the clothing-on-person mask $I_{tm}$, resulting in an overall matching loss expressed as follows:
\begin{equation}
\begin{aligned}
L_{Matching} & = \lambda_1 \cdot L_{1}(I_{\hat{c}}, I_{ct}) + \lambda_{reg} \cdot L_{reg} \\
& + \lambda_2 \cdot L_{1}(I_{\hat{cm}}, I_{tm}), 
\label{eq:loss_1}
\end{aligned}
\end{equation}
where $\lambda_1$, $\lambda_2$, and $\lambda_{reg}$ are set to $1$.

\begin{figure}[!t]
	\centering
	\includegraphics[width=1\linewidth]{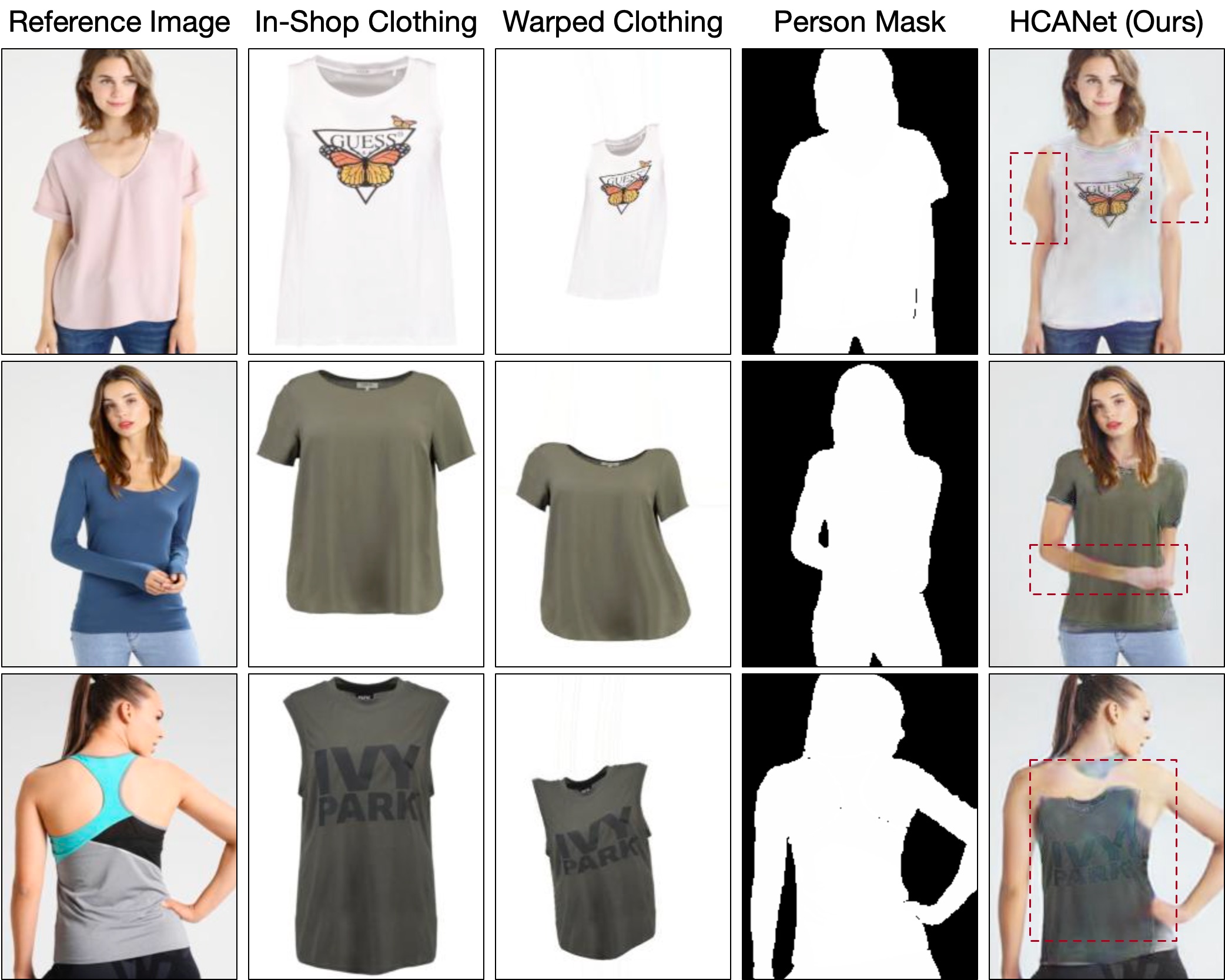} 
	\caption{Several failure cases of the proposed HCANet for virtual try-on.}
	\label{fig:failureCase}
\end{figure}

\begin{table*}[!t]
	\centering
	\caption{Ablation studies of the proposed HCANet.}
		\begin{tabular}{lcccc} \toprule
			Method &IoU $\uparrow$ & SSIM $\uparrow$ & LPIPS $\downarrow$ & IS $\uparrow$  \\ \hline	
			B0: Baseline~\cite{minar2020cp}  & 0.8425 & 0.8163 & 0.1144 & 3.1048 $\pm$ 0.1068 \\ 
			B1: B0 + HCA block for Stage I & 0.8541 & 0.8345 & 0.1103 & 3.1476 $\pm$ 0.0892 \\
			B2: B0 + HCA block for Stage II & 0.8587 & 0.8362 & 0.1097 & 3.1589 $\pm$ 0.0958 \\
			B3: B1 + B2 & 0.8624 & 0.8545 & 0.1046 & 3.2218 $\pm$ 0.0893 \\
			B4: B3 + $L_{1}$ mask loss & \textbf{0.8662} & \textbf{0.8591} & \textbf{0.1020} & \textbf{3.2483 $\pm$ 0.0865}  \\
			\bottomrule	
	\end{tabular}
	\label{tab:ablation}
\end{table*}

\begin{table}[!t]
	\centering
	\caption{Effect of feature fusion within HCA blocks.}
			\resizebox{1\linewidth}{!}{
		\begin{tabular}{cccccc} \toprule
			Stage I & Stage II &IoU $\uparrow$ & SSIM $\uparrow$ & LPIPS $\downarrow$ & IS $\uparrow$  \\ \hline	
            \ding{51} & \ding{55} & 0.8532 & 0.8329  & 0.1103 & 3.1428 $\pm$ 0.0965 \\
            \ding{55} & \ding{51} & 0.8611 & 0.8367 & 0.1084 & 3.1552 $\pm$ 0.0886 \\
			\ding{51} & \ding{51} & \textbf{0.8662} & \textbf{0.8591} & \textbf{0.1020} & \textbf{3.2483 $\pm$ 0.0865}  \\
			\bottomrule	
	\end{tabular}}
	\label{tab:fusion}
\end{table}

\begin{table}[!t]
	\centering
	\caption{Effect of the VGG loss.}
			\resizebox{1\linewidth}{!}{
		\begin{tabular}{ccccc} \toprule
			VGG loss &IoU $\uparrow$ & SSIM $\uparrow$ & LPIPS $\downarrow$ & IS $\uparrow$  \\ \hline	
            \ding{55} & 0.8572 & 0.8465 & 0.1105 & 3.1658 $\pm$ 0.0967 \\
		\ding{51} & \textbf{0.8662} & \textbf{0.8591} & \textbf{0.1020} & \textbf{3.2483 $\pm$ 0.0865}  \\
			\bottomrule	
	\end{tabular}}
	\label{tab:vgg}
\end{table}

\noindent\textbf{Ablation Analysis.}
The quantitative results for the HCA block variants are presented in Table~\ref{tab:ablation}.
Comparing the baseline B0 with B1, it is evident that B1 outperforms B0 significantly across all four metrics. This advantage can be attributed to the powerful long-range global dependence modeling capability of the proposed HCA block.
When we compare B0 with B2, adopting the HCA block only in the second try-on stage while keeping other settings the same as in \cite{minar2020cp}, we observe significant improvements across all four metrics.
Furthermore, combining B1 and B2 to create B3 results in further improvements in all metrics.
B4 represents the final model used in this paper, and it outperforms B3 by producing not only more naturally warped clothing but also more realistic try-on results.

\noindent\textbf{Effect of Feature Fusion within HCA Blocks.}
In HCA blocks, three separate feature maps (RGB, joints, and binary mask) are fused twice via cross-attention in both stages. By doing so, the model can learn richer and more informative representations and ultimately improve its performance.
In this section, we also provide empirical validation to support our claims about the effectiveness of feature fusion in the HCA blocks. We analyze the impact of feature fusion at different stages of our model, as detailed in Table \ref{tab:fusion}. Specifically, we evaluate three scenarios: (1) using feature fusion only in stage I, (2) using feature fusion only in stage II, and (3) using feature fusion in both stages. The results clearly show that incorporating feature fusion in both stages yields better performance compared to using it in only one stage. This demonstrates that feature fusion in both stages does indeed lead to richer and more informative representations, ultimately improving the model's overall performance.

\noindent\textbf{Effect of The VGG Loss.}
To validate the effectiveness of the VGG loss, we conduct an ablation study comparing the performance of our model with and without the VGG loss in Table \ref{tab:vgg}. The results clearly indicate that the model's performance decreases when the VGG loss is omitted. This demonstrates the critical role that VGG loss plays in enhancing the quality of the generated outputs, thereby justifying its inclusion in our loss function design.

\noindent\textbf{Hyper-Parameter Analysis.}
We investigate the influence of $\lambda_{1}$, $\lambda_{reg}$, $\lambda_{2}$ in Eq. \eqref{eq:loss_1} to the performance of our model. 
The results are shown in Table~\ref{tab:hyper}.
We see that when $\lambda_1$, $\lambda_2$, and $\lambda_{reg}$ are set to $1$, our model can achieve the best performance.

\begin{table}[!tbp]\small
	\centering
	\caption{The influence of $\lambda_{1}$, $\lambda_{reg}$, $\lambda_{2}$ in Eq. \eqref{eq:loss_1} to the performance of our model.}
	\begin{tabular}{cccc} \toprule
		$\lambda_{1}$        & $\lambda_{reg}$      & $\lambda_{2}$& FID $\downarrow$ \\ \midrule
          0.1 & 0.1 & 0.1 & 8.79 \\
          0.5 & 0.5 & 0.5 & 8.57 \\
          0.5 & 1 & 1 & 8.41 \\ 
          1 & 0.5 & 1 & 8.30 \\ 
          1 & 1 & 0.5 & 8.36 \\ 
		 1 & 1 & 1 & \textbf{8.25} \\ 
          5 & 1 & 1 & 8.47 \\ 
          1 & 5 & 1 & 8.42 \\ 
          1 & 1 & 5 & 8.56 \\ 
		 5 & 5 & 5 & 8.92 \\ 
          10 & 10 & 10 & 9.45 \\
          \bottomrule
	\end{tabular}
	\label{tab:hyper}
	\vspace{-0.4cm}
\end{table}

\subsection{Failure Cases and Analysis}
While HCANet produces impressive person try-on results, there are still inevitable failure cases. We identified three typical scenarios and provided visualizations in Figure~\ref{fig:failureCase} to illustrate them.
In the first case (first row), a substantial difference between the clothing in the reference image and the in-shop clothing image makes it challenging for the person's mask to match the new in-shop clothing effectively.
Self-occlusion is a significant issue in the second case (second row), resulting in the generation of ambiguous images.
In the third case (last row), very large poses and shape transformations can also lead to ambiguous results.
The primary cause of the first two failure cases is the lack of information in the input data about whether a body region should be covered with clothing. To address this, more accurate segmentation maps or finer-grained human annotations could be used.
As for the last case, 2D image-based methods struggle to completely capture large pose and shape transformations. Utilizing 3D input data, such as body meshes and 3D clothing models, may offer a solution to this issue.

%% file: 5conclusions.tex
\section{Conclusion}

In addressing the intricate challenges posed by the virtual try-on task, we introduce our innovative solution, the Hierarchical Cross-Attention Network, denoted as HCANet.
HCANet is designed with a two-stage generative approach, leveraging the novel Hierarchical Cross-Attention (HCA) block. This strategic utilization of the HCA block enables the network to effectively capture intricate correlations between the person and clothing modalities. By hierarchically modeling the dependencies between these modalities, HCANet excels in preserving the nuanced details of the clothing items and their interaction with the wearer, leading to more realistic virtual try-on outcomes.
To comprehensively validate the efficacy of HCANet, we conducted a series of experiments encompassing both human assessment and automated evaluation metrics. The results of our experiments demonstrate that HCANet not only outperforms existing methods but also establishes new state-of-the-art benchmarks in the virtual try-on domain. The qualitative and quantitative assessments affirm the superior performance of HCANet in generating visually convincing and contextually accurate virtual try-on results, thus making a substantial contribution to the advancement of virtual try-on technologies.

Exploring how our model handles obstructions is indeed an intriguing avenue for future research. We intend to delve deeper into this aspect in our upcoming work, as obstructions can significantly impact the accuracy and realism of the virtual fitting experience. By addressing this challenge, we aim to enhance the overall capabilities of our model and provide users with even more reliable and lifelike try-on results.

\section{Acknowledgments}
This work was supported by the MUR PNRR project FAIR (PE00000013) funded by the NextGenerationEU, the PRIN project CREATIVE (Prot. 2020ZSL9F9), the Major Project of Natural Science Foundation of Jiangsu Education Department (no.22KJA630001), the National Natural Science Foundation of China (NSFC, no. 61701243), and the Fundamental Research Funds for the Central Universities, Peking University.

%% file: tmm.bbl
\begin{thebibliography}{10}
\providecommand{\url}[1]{#1}
\csname url@samestyle\endcsname
\providecommand{\newblock}{\relax}
\providecommand{\bibinfo}[2]{#2}
\providecommand{\BIBentrySTDinterwordspacing}{\spaceskip=0pt\relax}
\providecommand{\BIBentryALTinterwordstretchfactor}{4}
\providecommand{\BIBentryALTinterwordspacing}{\spaceskip=\fontdimen2\font plus
\BIBentryALTinterwordstretchfactor\fontdimen3\font minus
  \fontdimen4\font\relax}
\providecommand{\BIBforeignlanguage}[2]{{%
\expandafter\ifx\csname l@#1\endcsname\relax
\typeout{** WARNING: IEEEtran.bst: No hyphenation pattern has been}%
\typeout{** loaded for the language `#1'. Using the pattern for}%
\typeout{** the default language instead.}%
\else
\language=\csname l@#1\endcsname
\fi
#2}}
\providecommand{\BIBdecl}{\relax}
\BIBdecl

\bibitem{han2018viton}
X.~Han, Z.~Wu, Z.~Wu, R.~Yu, and L.~S. Davis, ``Viton: An image-based virtual
  try-on network,'' in \emph{CVPR}, 2018.

\bibitem{jetchev2017conditional}
N.~Jetchev and U.~Bergmann, ``The conditional analogy gan: Swapping fashion
  articles on people images,'' in \emph{ICCV Workshops}, 2017.

\bibitem{wang2018toward}
B.~Wang, H.~Zheng, X.~Liang, Y.~Chen, L.~Lin, and M.~Yang, ``Toward
  characteristic-preserving image-based virtual try-on network,'' in
  \emph{ECCV}, 2018.

\bibitem{minar2020cp}
M.~Minar, T.~Tuan, H.~Ahn, P.~Rosin, and Y.~Lai, ``Cp-vton+: Clothing shape and
  texture preserving image-based virtual try-on,'' in \emph{CVPR Workshops},
  2020.

\bibitem{ehara2006texture}
J.~Ehara and H.~Saito, ``Texture overlay for virtual clothing based on pca of
  silhouettes,'' in \emph{ACM ISMAR}, 2006.

\bibitem{brouet2012design}
R.~Brouet, A.~Sheffer, L.~Boissieux, and M.-P. Cani, ``Design preserving
  garment transfer,'' \emph{ACM TOG}, vol.~31, no.~4, 2012.

\bibitem{chen2016synthesizing}
W.~Chen, H.~Wang, Y.~Li, H.~Su, Z.~Wang, C.~Tu, D.~Lischinski, D.~Cohen-Or, and
  B.~Chen, ``Synthesizing training images for boosting human 3d pose
  estimation,'' in \emph{3DV}, 2016.

\bibitem{guan2012drape}
P.~Guan, L.~Reiss, D.~A. Hirshberg, A.~Weiss, and M.~J. Black, ``Drape:
  Dressing any person,'' \emph{ACM TOG}, vol.~31, no.~4, pp. 1--10, 2012.

\bibitem{sekine2014virtual}
M.~Sekine, K.~Sugita, F.~Perbet, B.~Stenger, and M.~Nishiyama, ``Virtual
  fitting by single-shot body shape estimation,'' in \emph{3D Body Scanning
  Technologies}, 2014.

\bibitem{ge2021disentangled}
C.~Ge, Y.~Song, Y.~Ge, H.~Yang, W.~Liu, and P.~Luo, ``Disentangled cycle
  consistency for highly-realistic virtual try-on,'' in \emph{CVPR}, 2021.

\bibitem{yang2020towards}
H.~Yang, R.~Zhang, X.~Guo, W.~Liu, W.~Zuo, and P.~Luo, ``Towards
  photo-realistic virtual try-on by adaptively generating-preserving image
  content,'' in \emph{CVPR}, 2020.

\bibitem{ge2021parser}
Y.~Ge, Y.~Song, R.~Zhang, C.~Ge, W.~Liu, and P.~Luo, ``Parser-free virtual
  try-on via distilling appearance flows,'' in \emph{CVPR}, 2021.

\bibitem{lee2022towards}
S.~Lee, S.~Lee, and J.~Lee, ``Towards detailed characteristic-preserving
  virtual try-on,'' in \emph{CVPR Workshops}, 2022.

\bibitem{fenocchi2022dual}
E.~Fenocchi, D.~Morelli, M.~Cornia, L.~Baraldi, F.~Cesari, and R.~Cucchiara,
  ``Dual-branch collaborative transformer for virtual try-on,'' in \emph{CVPR
  Workshops}, 2022.

\bibitem{he2022style}
S.~He, Y.-Z. Song, and T.~Xiang, ``Style-based global appearance flow for
  virtual try-on,'' in \emph{CVPR}, 2022.

\bibitem{huang2022towards}
Z.~Huang, H.~Li, Z.~Xie, M.~Kampffmeyer, X.~Liang \emph{et~al.}, ``Towards
  hard-pose virtual try-on via 3d-aware global correspondence learning,''
  \emph{NeurIPS}, 2022.

\bibitem{fele2022c}
B.~Fele, A.~Lampe, P.~Peer, and V.~Struc, ``C-vton: Context-driven image-based
  virtual try-on network,'' in \emph{WACV}, 2022.

\bibitem{morelli2022dress}
D.~Morelli, M.~Fincato, M.~Cornia, F.~Landi, F.~Cesari, and R.~Cucchiara,
  ``Dress code: High-resolution multi-category virtual try-on,'' in
  \emph{CVPR}, 2022.

\bibitem{bai2022single}
S.~Bai, H.~Zhou, Z.~Li, C.~Zhou, and H.~Yang, ``Single stage virtual try-on via
  deformable attention flows,'' in \emph{ECCV}, 2022.

\bibitem{ren2023cloth}
B.~Ren, H.~Tang, F.~Meng, D.~Runwei, P.~H. Torr, and N.~Sebe, ``Cloth
  interactive transformer for virtual try-on,'' \emph{ACM TOMM}, vol.~20,
  no.~4, pp. 1--20, 2023.

\bibitem{belongie2002shape}
S.~Belongie, J.~Malik, and J.~Puzicha, ``Shape matching and object recognition
  using shape contexts,'' \emph{IEEE TPAMI}, vol.~24, no.~4, pp. 509--522,
  2002.

\bibitem{bookstein1989principal}
F.~L. Bookstein, ``Principal warps: Thin-plate splines and the decomposition of
  deformations,'' \emph{IEEE TPAMI}, vol.~11, no.~6, pp. 567--585, 1989.

\bibitem{rocco2017convolutional}
I.~Rocco, R.~Arandjelovic, and J.~Sivic, ``Convolutional neural network
  architecture for geometric matching,'' in \emph{CVPR}, 2017.

\bibitem{yu2018generative}
J.~Yu, Z.~Lin, J.~Yang, X.~Shen, X.~Lu, and T.~S. Huang, ``Generative image
  inpainting with contextual attention,'' in \emph{CVPR}, 2018.

\bibitem{wu2023learning}
J.~Wu, H.~Liu, Y.~Su, W.~Shi, and H.~Tang, ``Learning concordant attention via
  target-aware alignment for visible-infrared person re-identification,'' in
  \emph{ICCV}, 2023.

\bibitem{zhou2022cloth}
Z.~Zhou, H.~Liu, W.~Shi, H.~Tang, and X.~Shi, ``A cloth-irrelevant harmonious
  attention network for cloth-changing person re-identification,'' in
  \emph{ICPR}, 2022.

\bibitem{ren2023transductive}
H.~Ren, S.~Liu, X.~Yu, L.~Zou, Y.~Zhou, X.~Wang, and H.~Tang, ``Transductive
  prototypical attention reasoning network for few-shot sar target
  recognition,'' \emph{IEEE TGRS}, 2023.

\bibitem{anwar2019real}
S.~Anwar and N.~Barnes, ``Real image denoising with feature attention,'' in
  \emph{CVPR}, 2019.

\bibitem{xu2018structured}
D.~Xu, W.~Wang, H.~Tang, H.~Liu, N.~Sebe, and E.~Ricci, ``Structured attention
  guided convolutional neural fields for monocular depth estimation,'' in
  \emph{CVPR}, 2018.

\bibitem{yang2021transformer}
G.~Yang, H.~Tang, M.~Ding, N.~Sebe, and E.~Ricci, ``Transformer-based attention
  networks for continuous pixel-wise prediction,'' in \emph{ICCV}, 2021.

\bibitem{tang2019multi}
H.~Tang, D.~Xu, N.~Sebe, Y.~Wang, J.~J. Corso, and Y.~Yan, ``Multi-channel
  attention selection gan with cascaded semantic guidance for cross-view image
  translation,'' in \emph{CVPR}, 2019.

\bibitem{tang2021attentiongan}
H.~Tang, H.~Liu, D.~Xu, P.~H. Torr, and N.~Sebe, ``Attentiongan: Unpaired
  image-to-image translation using attention-guided generative adversarial
  networks,'' \emph{IEEE TNNLS}, 2021.

\bibitem{tang2019attention}
H.~Tang, D.~Xu, N.~Sebe, and Y.~Yan, ``Attention-guided generative adversarial
  networks for unsupervised image-to-image translation,'' in \emph{IJCNN},
  2019.

\bibitem{tang2020dual}
H.~Tang, S.~Bai, and N.~Sebe, ``Dual attention gans for semantic image
  synthesis,'' in \emph{ACM MM}, 2020.

\bibitem{tang2019attribute}
H.~Tang, X.~Chen, W.~Wang, D.~Xu, J.~J. Corso, N.~Sebe, and Y.~Yan,
  ``Attribute-guided sketch generation,'' in \emph{FG}, 2019.

\bibitem{tang2020xinggan}
H.~Tang, S.~Bai, L.~Zhang, P.~H. Torr, and N.~Sebe, ``Xinggan for person image
  generation,'' in \emph{ECCV}, 2020.

\bibitem{tang2020edge}
H.~Tang, X.~Qi, G.~Sun, D.~Xu, N.~Sebe, R.~Timofte, and L.~Van~Gool, ``Edge
  guided gans with contrastive learning for semantic image synthesis,'' in
  \emph{ICLR}, 2023.

\bibitem{wu2022cross}
S.~Wu, H.~Tang, X.-Y. Jing, J.~Qian, N.~Sebe, Y.~Yan, and Q.~Zhang,
  ``Cross-view panorama image synthesis with progressive attention gans,''
  \emph{Elsevier PR}, 2022.

\bibitem{tang2022multi}
H.~Tang, P.~H. Torr, and N.~Sebe, ``Multi-channel attention selection gans for
  guided image-to-image translation,'' \emph{IEEE TPAMI}, vol.~45, no.~5, pp.
  6055--6071, 2022.

\bibitem{tang2023graph}
H.~Tang, Z.~Zhang, H.~Shi, B.~Li, L.~Shao, N.~Sebe, R.~Timofte, and
  L.~Van~Gool, ``Graph transformer gans for graph-constrained house
  generation,'' in \emph{CVPR}, 2023.

\bibitem{duan2021cascade}
B.~Duan, W.~Wang, H.~Tang, H.~Latapie, and Y.~Yan, ``Cascade attention guided
  residue learning gan for cross-modal translation,'' in \emph{ICPR}, 2021.

\bibitem{duan2021audio}
B.~Duan, H.~Tang, W.~Wang, Z.~Zong, G.~Yang, and Y.~Yan, ``Audio-visual event
  localization via recursive fusion by joint co-attention,'' in \emph{WACV},
  2021.

\bibitem{ng2020solar}
T.~Ng, V.~Balntas, Y.~Tian, and K.~Mikolajczyk, ``Solar: second-order loss and
  attention for image retrieval,'' in \emph{ECCV}, 2020.

\bibitem{hu2022supervised}
J.~Hu, X.~Zhi, S.~Jiang, H.~Tang, W.~Zhang, and L.~Bruzzone, ``Supervised
  multi-scale attention-guided ship detection in optical remote sensing
  images,'' \emph{IEEE TGRS}, vol.~60, pp. 1--14, 2022.

\bibitem{ding2023few}
H.~Ding, C.~Sun, H.~Tang, D.~Cai, and Y.~Yan, ``Few-shot medical image
  segmentation with cycle-resemblance attention,'' in \emph{WACV}, 2023.

\bibitem{ding2020lanet}
L.~Ding, H.~Tang, and L.~Bruzzone, ``Lanet: Local attention embedding to
  improve the semantic segmentation of remote sensing images,'' \emph{IEEE
  TGRS}, vol.~59, no.~1, pp. 426--435, 2020.

\bibitem{yang2022continual}
G.~Yang, E.~Fini, D.~Xu, P.~Rota, M.~Ding, T.~Hao, X.~Alameda-Pineda, and
  E.~Ricci, ``Continual attentive fusion for incremental learning in semantic
  segmentation,'' \emph{IEEE TMM}, 2022.

\bibitem{wang2020axial}
H.~Wang, Y.~Zhu, B.~Green, H.~Adam, A.~Yuille, and L.-C. Chen, ``Axial-deeplab:
  Stand-alone axial-attention for panoptic segmentation,'' in \emph{ECCV},
  2020.

\bibitem{johnson2016perceptual}
J.~Johnson, A.~Alahi, and L.~Fei-Fei, ``Perceptual losses for real-time style
  transfer and super-resolution,'' in \emph{ECCV}, 2016.

\bibitem{wang2004image}
Z.~Wang, A.~C. Bovik, H.~R. Sheikh, and E.~P. Simoncelli, ``Image quality
  assessment: from error visibility to structural similarity,'' \emph{IEEE
  TIP}, vol.~13, no.~4, pp. 600--612, 2004.

\bibitem{zhang2018unreasonable}
R.~Zhang, P.~Isola, A.~A. Efros, E.~Shechtman, and O.~Wang, ``The unreasonable
  effectiveness of deep features as a perceptual metric,'' in \emph{CVPR},
  2018.

\bibitem{salimans2016improved}
T.~Salimans, I.~Goodfellow, W.~Zaremba, V.~Cheung, A.~Radford, and X.~Chen,
  ``Improved techniques for training gans,'' in \emph{NeurIPS}, 2016.

\bibitem{heusel2017gans}
M.~Heusel, H.~Ramsauer, T.~Unterthiner, B.~Nessler, and S.~Hochreiter, ``Gans
  trained by a two time-scale update rule converge to a local nash
  equilibrium,'' in \emph{NeurIPS}, 2017.

\bibitem{han2019clothflow}
X.~Han, X.~Hu, W.~Huang, and M.~R. Scott, ``Clothflow: A flow-based model for
  clothed person generation,'' in \emph{ICCV}, 2019.

\bibitem{chopra2021zflow}
A.~Chopra, R.~Jain, M.~Hemani, and B.~Krishnamurthy, ``Zflow: Gated appearance
  flow-based virtual try-on with 3d priors,'' in \emph{ICCV}, 2021.

\bibitem{du2024greatness}
C.~Du, S.~Liu, S.~Xiong \emph{et~al.}, ``Greatness in simplicity: Unified
  self-cycle consistency for parser-free virtual try-on,'' \emph{NeurIPS},
  2023.

\bibitem{yang2023occlumix}
Z.~Yang, J.~Chen, Y.~Shi, H.~Li, T.~Chen, and L.~Lin, ``Occlumix: Towards
  de-occlusion virtual try-on by semantically-guided mixup,'' \emph{IEEE TMM},
  2023.

\end{thebibliography}
